\documentclass[10pt,journal,compsoc]{IEEEtran}

\usepackage[numbers]{natbib}
\usepackage{amsmath,amssymb,amsfonts}
\usepackage{algpseudocode}
\usepackage{algorithm}
\usepackage{graphicx}
\usepackage{textcomp}
\usepackage{xcolor}
\usepackage{multirow}
\usepackage{subcaption}
\usepackage{color}
\usepackage{xspace}
\usepackage{lipsum}
\usepackage{amsmath}
\usepackage{bm}
\usepackage{xcolor}

\DeclareMathOperator*{\argmax}{argmax} 

\begin{document}

\title{Tackling Virtual and Real Concept Drifts: An Adaptive Gaussian Mixture Model Approach\let\footnote\thanks\footnote{A preliminary version of this research was published in \cite{oliveira2019gmm}.}}

\author{Gustavo~Oliveira,
        Leandro~Minku,~\IEEEmembership{Member,~IEEE,}
        and~Adriano~Oliveira,~\IEEEmembership{Senior Member,~IEEE,}

\IEEEcompsocitemizethanks{

\IEEEcompsocthanksitem G. Oliveira and A. Oliveira were with the Centro de Inform\'atica, Universidade Federal de Pernambuco, Recife, Pernambuco, Brazil.\protect\\
E-mail: ghfmo@cin.ufpe.br and alio@cin.ufpe.br.
\IEEEcompsocthanksitem L. Minku was with the School of Computer Science, University of Birmingham, Birmingham, UK.\protect\\
E-mail: L.L.Minku@cs.bham.ac.uk.
}

\thanks{Manuscript received April 19, 200X; revised August 26, 201X.}}

\markboth{Journal of \LaTeX\ Class Files,~Vol.~14, No.~X, August~201X}%
{Shell \MakeLowercase{\textit{et al.}}: Bare Demo of IEEEtran.cls for Computer Society Journals}

\IEEEtitleabstractindextext{
\begin{abstract}
Real-world applications have been dealing with large amounts of data that arrive over time and generally present changes in their underlying joint probability distribution, i.e., concept drift. Concept drift can be subdivided into two types: virtual drift, which affects the unconditional probability distribution $p(\bm{x})$, and real drift, which affects the conditional probability distribution $p(y|\bm{x})$. Existing works focuses on real drift. However, strategies to cope with real drift may not be the best suited for dealing with virtual drift, since the real class boundaries remain unchanged. We provide the first in depth analysis of the differences between the impact of virtual and real drifts on classifiers' suitability. We propose an approach to handle both drifts called \textcolor{black}{On-line} Gaussian Mixture Model With Noise Filter For Handling Virtual and Real Concept Drifts (OGMMF-VRD). Experiments with 7 synthetic and 3 real-world datasets show that OGMMF-VRD obtained the best results in terms of average accuracy, G-mean and runtime compared to existing approaches. Moreover, its accuracy over time suffered less performance degradation in the presence of drifts.
\end{abstract}

}

\maketitle

\IEEEdisplaynontitleabstractindextext

\IEEEpeerreviewmaketitle

\IEEEraisesectionheading{\section{Introduction}\label{sec:introduction}}

In recent years, real-world applications like credit card fraud detection, flight delay and weather forecasting have been dealing with tremendous growth in the amount of data, which typically arrive continuously and sequentially over time and evolve due to the underlying dynamics of real-world activities. Such sequences of data are known as data streams \textcolor{black}{\cite{oliveira2017time, song2019fuzzy}}. They are challenging for data modeling systems \cite{gama2014survey}, requiring classifiers to adapt to changes over time. Changes in the underlying distributions of the problem are called concept drifts \cite{minku2012ddd}.


Concept drift can be subdivided into two types: virtual drift and real drift \cite{gama2014survey}. Virtual drift can be defined as a change in the unconditional probability distribution $P(\textbf{x})$ and real drift can be defined as a change in the conditional probabilities $P(y|\textbf{x})$. They may occur separately or simultaneously and may have different impacts on the classifier performance.

Most existing work on data stream learning focuses on real drifts, because such drifts change the \textit{true} decision boundaries of the problem, directly degrading the performance of classifiers \cite{khamassi2018discussion}. As virtual drifts do not change the \textit{true} decision boundaries of the problem, they attracted much less attention from the research community. Nevertheless, they can also affect the classifier performance, because they may affect the suitability of the decision boundaries \textit{learned} by the classifiers. For instance, the appearance of observations in regions of the space that were not covered by training examples may reveal insufficient or incorrectly \textit{learned} decision boundaries, which need to be adjusted for the classifier to remain suitable. 

No existing work provides an in depth understanding of the differences between the effect of these two types of drift on the suitability of classifiers. As a result, existing data stream learning approaches treat virtual drifts using the same strategies as for real drifts \cite{khamassi2018discussion}. A common strategy to adapt to a new (previously unseen) concept is to create a new classifier to learn it. However, such strategy may not be the best for dealing with virtual drifts. This is because the knowledge acquired before the drift may remain valid after a virtual drift occurs \cite{gama2014survey}, given that the \textit{true} decision boundaries do not change. Learning a new classifier from scratch thus wastes potentially useful knowledge that could speed up adaptation to virtual drifts.



Moreover, the strategy of creating new models can be prone to noise, which could be very problematic in the presence of virtual drifts. For instance, approaches based on drift detectors could potentially be tuned to detect minor changes in the underlying distribution such as virtual drifts. This tuning may cause the system to confuse these drifts with noise, thus triggering the unnecessary creation of new models. This could potentially harm system predictive performance as a whole, given that new models require incoming observations to train to become accurate.



With that in mind, this paper provides the first in-depth analysis of the differences between the impact caused by virtual and real drifts on the suitability of approaches based on bayes theorem, as they lend themselves to dealing with different types of drift, due to their pertinence inferences \cite{oliveira2019gmm}. Besides that, we propose a new approach to handle both virtual and real drifts simultaneously while achieving more robustness to noise. This approach has been guided by the following Research Questions (RQs), which have not been considered by previous work:

\textbf{RQ1) What is the difference between the impact of virtual and real drifts on the suitability of classifiers' \textit{learned} decision boundaries and predictive performance over time?} \label{qt:one} This RQ provides the foundation for proposing novel approaches able to more efficiently and effectively deal with both types of drift at the same time. We hypothesize that when virtual drifts occurs, the previously \textit{learned} decision boundaries remain suitable, and the only thing that needs to be done is to learn the emerging region. We also hypothesized that when non-severe real drifts happen, only a small portion of the \textit{learned} decision boundaries becomes unsuitable, whereas severe real drifts require a significant reset of the \textit{learned} decision boundaries.


\textbf{RQ2) How to deal with both virtual and real drifts while achieving robustness to noise?} \label{qt:two} We explore the potential of GMM to enable different strategies to be used to tackle these different types of drift. GMM has pertinence inferences useful for virtual drifts, which enable us to verify whether or not a new observation belongs to the trained distribution. If new observations arrive in regions of the space that are not covered by existing Gaussians, a new Gaussian can be created for them. As incoming observations could be noise, it is essential for the work to handle it. Therefore, we hypothesize that creating a filter using techniques of instance hardness can help the system to be less affected by noise. Meanwhile, the model could update its existing Gaussians to cope with non-severe real drifts, and could still be restarted to deal with severe real drifts if its \textit{learned} decision boundaries become largely unsuitable. \label{hp:two}


\textbf{RQ3) How to best harness knowledge gained from past similar concepts to accelerate adaptation to both virtual and real drifts?} \label{qt:three} The most widely used strategy to deal with concept drift is to learn previously unseen concepts from scratch. This forces the system to use obsolete models until sufficient new data have arrived for retraining, causing a large degradation in performance. Some studies have considered the storage of past models in a pool to accelerate adaptation to recurrent concepts \cite{almeida2018adapting}. We hypothesize that saving past GMMs in a pool can not only help the system to adapt to recurrent concepts, but also to accelerate adaptation to virtual and real drifts that lead to new concepts that share similarities to old concepts. \label{hp:three}

  
In our previous work \cite{oliveira2019gmm}, we preliminary demonstrated the potential benefit of GMMs for tackling virtual and real drifts. However, that preliminary work (i) did not provide a detailed understanding of how virtual and real drifts affect classifier performances; (ii) ignored the impact of noisy observations on virtual drift adaptation, which makes the system to create Gaussians in unwanted regions which can cause misclassification; (iii) delayed the learning of new regions through Gaussians because it was limited only to very distant areas; (iv) delayed the ability to track data evolution since the Gaussians were updated only when misclassification occurred; and (v) in real drift adaptation, new concepts were learned entirely from scratch, which in turn discard useful knowledge that could be useful in the drift adaptation. Together, these drawbacks limited the predictive performance of that framework. Therefore, the current paper proposes a new approach called \textcolor{black}{On-line} Gaussian Mixture Model With Noise Filter For Handling Virtual and Real Concept Drifts (OGMMF-VRD), which overcomes these problems.


This paper is further organized as follows. Section \ref{sec:related work} explains related work. Section \ref{sec:problem definition} presents our problem formulation. Section \ref{subsec:datasets} presents the datasets used in our study. Section \ref{sec:drift impacts} presents our analysis to answer RQ1. Based on that, Section \ref{sec:proposed method} proposes our approach OGMMF-VRD, partly answering RQ2 and RQ3. Section \ref{sec:experiments} presents our experimental analysis of the proposed approach, completing the answer to RQ2 and RQ3. Section \ref{sec:conclusion} concludes the paper and gives directions for further research.

\begin{figure*}[t]
    \begin{subfigure}[h]{0.23\textwidth}
        \centering
        \includegraphics[height=1.35in]{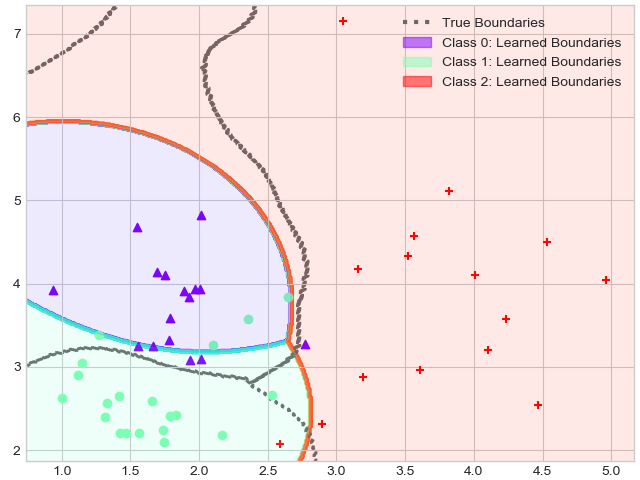}
        \caption{\textcolor{black}{Dynse - Virtual 9 (Bfr.)}}
        \label{fig:dynse virtual9 before}
    \end{subfigure}
    ~
    \begin{subfigure}[h]{0.23\textwidth}
        \centering
        \includegraphics[height=1.35in]{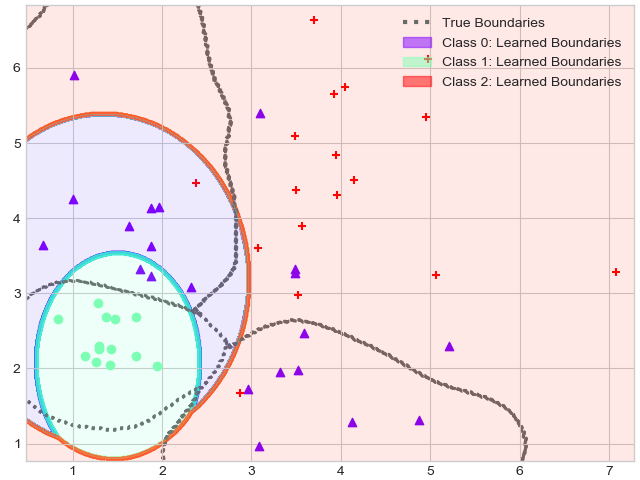}
        \caption{Dynse - Virtual 9 (Aft.)}
        \label{fig:dynse virtual9 after}
    \end{subfigure}
    ~
    \begin{subfigure}[h]{0.23\textwidth}
        \centering
        \includegraphics[height=1.35in]{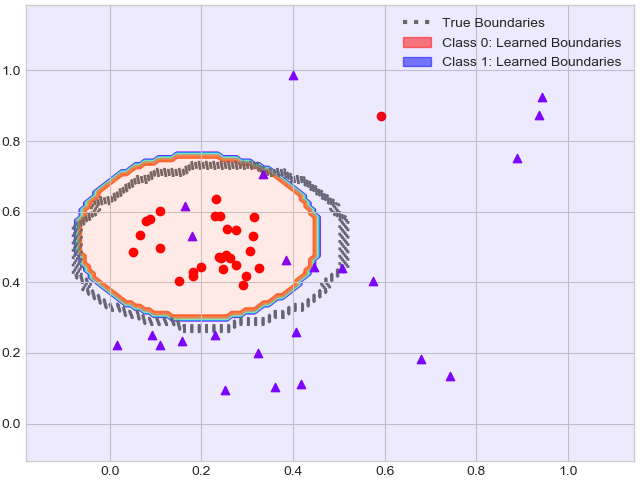}
        \caption{\textcolor{black}{Dynse - Circles (Bfr.)}}
        \label{fig:dynse circles before}
    \end{subfigure}
    ~
    \begin{subfigure}[h]{0.23\textwidth}
        \centering
        \includegraphics[height=1.35in]{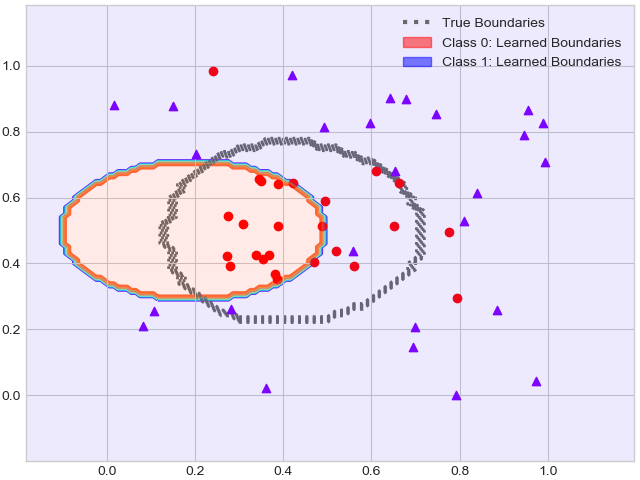}
        \caption{Dynse - Circles (Aft.)}
        \label{fig:dynse circles after}
    \end{subfigure}
    ~
    \begin{subfigure}[h]{0.241\textwidth}
        \centering
        \includegraphics[height=1.35in]{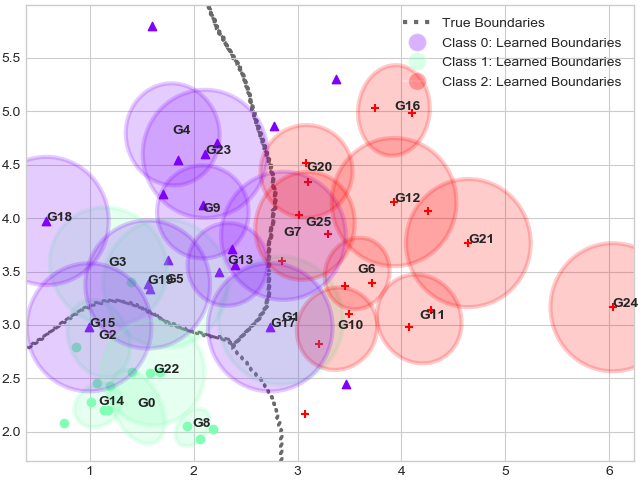}
        \caption{\textcolor{black}{IGMM-CD - Virtual 9 (Bfr.)}}
        \label{fig:igmmcd virtual9 before}
    \end{subfigure}
    ~
    \begin{subfigure}[h]{0.241\textwidth}
        \centering
        \includegraphics[height=1.35in]{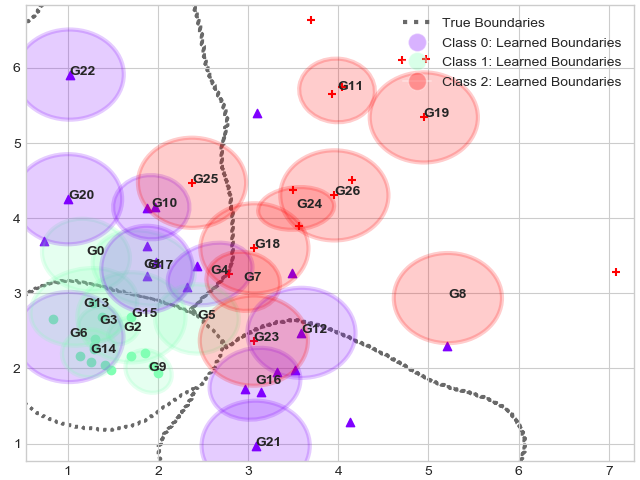}
        \caption{IGMM-CD - Virtual 9 (Aft.)}
        \label{fig:igmmcd virtual9 after}
    \end{subfigure}
    ~
    \begin{subfigure}[h]{0.241\textwidth}
        \centering
        \includegraphics[height=1.35in]{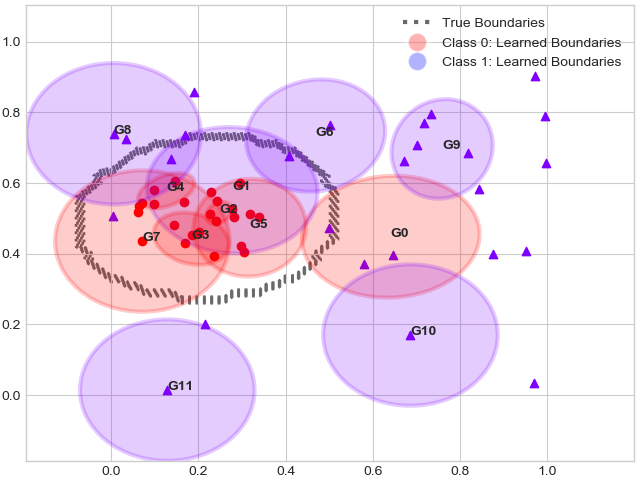}
        \caption{\textcolor{black}{IGMM-CD - Circles (Bfr.)}}
        \label{fig:igmmcd circles before}
    \end{subfigure}
    ~
    \begin{subfigure}[h]{0.23\textwidth}
        \centering
        \includegraphics[height=1.35in]{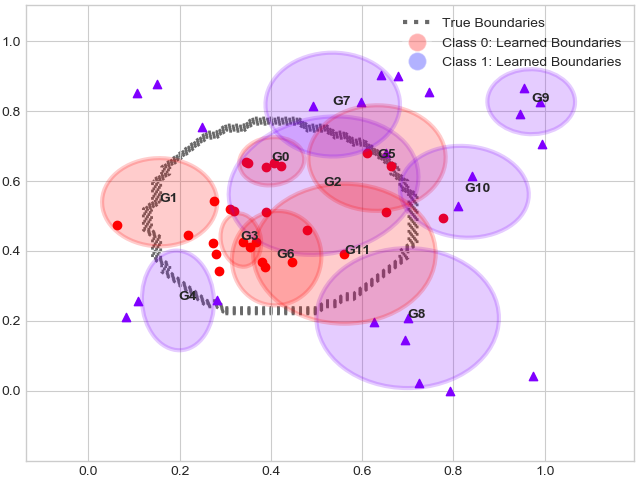}
        \caption{IGMM-CD - Circles (Aft.)}
        \label{fig:igmmcd circles after}
    \end{subfigure}
    \caption{Execution of Dynse \cite{almeida2018adapting} and IGMM-CD \cite{oliveira2015igmm} on Virtual 9 and Circles datasets (Tbl. \ref{tbl:datasets}). The points represent the first 50 \textcolor{black}{before (bfr.) and after (aft.)} the concept drift. Dynse uses as base classifier the Gaussian Naive Bayes and IGMM-CD the GMM. Gray and dotted lines represent the \textit{true} decision boundaries of the problem. The solid colored lines represent the decision boundaries \textit{learned} by each approach. The text G0, ..Gn represents the Gaussian number.}
    \label{fig:concept drift impacts}
\end{figure*}

\section{Related Work}
\label{sec:related work}

In general, several existing methods have been proposed to deal with concept drift, and a few different surveys discuss these approaches. For general learning in non-stationary environments, we refer readers to \cite{ditzler2015learning, gama2014survey}. For ensemble approaches for non-stationary environments, we refer readers to \cite{krawczyk2017ensemble,gomes2017survey}. For class imbalance learning in non-stationary environments, we refer readers to \cite{wang2018systematic}. Despite all these studies, few of the existing approaches differentiate between virtual and real drifts, and propose to handle both. In this sense, the following works stand out:


Oliveira et al.'s Incremental Gaussian Mixture Model for Concept Drift (IGMM-CD) \cite{oliveira2015igmm} uses \textit{on-line} learning to incorporate new incoming observations. If an incoming observation is classified correctly, the nearest Gaussian to it is updated in an attempt to handle virtual drifts. If the system misclassifies the incoming observation and does not have a Gaussian with the minimum distance ($Cver$) to it, a new Gaussian with size ($sigma\_ini$) is created in an attempt to handle real drifts. Besides that, the system excludes Gaussians with lower density if the predefined number of Gaussians per class ($T$) is exceeded. Despite that, virtual drifts resulting in misclassifications are treated in the same way as real drifts, causing the unnecessary addition of new Gaussians. In addition, this approach suffers from delays in excluding obsolete Gaussians in the presence of abrupt drifts, degrading performance.

        
Almeida et al.'s Dynamic Selection Based Drift Handler (Dynse) is an approach based on Dynamic Classifier Selection (DCS) \cite{almeida2018adapting}. DCS is the process of selecting a specific classifier for each test instance according to its neighborhood ($k$) in a validation set. The validation set ($M$) used by Dynse is represented by a sliding window which traverses the incoming data excluding the oldest observations. Virtual drifts can be dealt with by using a sliding window with very large size, because it will contain many observations corresponding to the current concept. Real drifts can be dealt with by using a sliding window with small size, since its observations can be excluded faster. New classifiers are added into a pool ($D$) at each batch of $m$ observations to learn new concepts as they arrive. However, as Dynse relies on a pre-defined window size, it can only deal with one type of drift (virtual or real) at a given run. If the data stream presents both virtual and real drifts, or if the chosen window size does not match the type of drift presented in the data stream, Dynse is likely to have its predictive performance negatively affected.


Oliveira et al.'s Gaussian Mixture Model for dealing with Virtual and Real concept Drifts (GMM-VRD) \cite{oliveira2019gmm} combines \textit{batch} and \textit{on-line} learning to handle both virtual and real drifts. A GMM is trained using a batch with $m$ observations. The parameter ($Kmax$) is used to define the maximum number of Gaussians and the Akaike Information Criterion (AIC) is used to define the best number. If a misclassification occurs, the pertinence of the new incoming observation is used to decide whether to update or to create new Gaussians to handle virtual drifts. In parallel, a Concept Drift Test (CDT) is used to monitor the system's performance. If the performance is decreasing, the system is reset to cope with real drifts. The drawback of this approach is its sensitivity to noisy observations, which can cause creation and update of Gaussians on unwanted regions, degrading performance. 


Our approach OGMMF-VRD is proposed to overcome the above mentioned problems of existing approaches, aiming at dealing with both virtual and real drifts concurrently.




\section{Problem Definition}
\label{sec:problem definition}

Consider a data stream \textcolor{black}{as follows: $S = \{(\textbf{x}_1,y_1), (\textbf{x}_2,y_2), \cdots, (\textbf{x}_t,y_t), \cdots \}$,} where $\textbf{x}_t \in \chi$ is a $d$-dimensional vector of input attributes, $y_t \in \gamma$ is a categorical output attribute, $\chi$ is the input space and $\gamma$ is the output space, in which each observation $(\textbf{x}_t,y_t)$ comes from a joint probability distribution $P_{t}(\textbf{x}, y)$.

On-line supervised learning from this kind of data consists of creating a model $f_{t}: \chi \rightarrow \gamma$ where at each new time step $t$, the previous model $f_{t-1}$ is updated with the new incoming observation $(\textbf{x}_t,y_t)$ to be able to generalize to unseen observations of $P_{t}(\textbf{x}, y)$. 

A challenge faced by on-line supervised learning is that observations produced at distinct time steps $t$ and $t + \Delta$ may come from different joint probability  distributions $P_{t}(\textbf{x}, y) \neq P_{t + \Delta}(\textbf{x}, y)$, i.e., the data stream may present concept drift \cite{webb2016characterizing}. 
%
%
%
The joint probability is formalized as $P_t(\textbf{x}, y) = P_t(y|\textbf{x})P_t(\textbf{x})$, where $P_t(\textbf{x})$ is the probability distribution of inputs and $P_t(y|\textbf{x})$ is the conditional probability of the outputs given the inputs. The latter represents the \textit{true} decision boundaries of the problem. So, drifts can happen if $P_{t}(\textbf{x}) \neq P_{t + \Delta}(\textbf{x})$, if $P_{t}(y|\textbf{x}) \neq P_{t + \Delta}(y|\textbf{x})$, or both. Drifts affecting $P(\textbf{x})$ are categorized as virtual drifts and drifts affecting $P(y|\textbf{x})$ are categorized as real drifts \cite{yamauchi2010incremental}. Only real drifts affect the \textit{true} decision boundaries of the problem, whereas both types of drift may affect the suitability of the \textit{learned} decision boundaries, depending on the learning algorithm and model being adopted.


\section{Datasets}
\label{subsec:datasets}

We used synthetic and real-world datasets, whose characteristics are presented in Tbl. \ref{tbl:datasets}. \textcolor{black}{These datasets are available on Github}\footnote{https://github.com/GustavoHFMO/OGMMF-VRD\_datasets}. All datasets described in Tbl. \ref{tbl:datasets} have continuous attributes, \textcolor{black}{as data clustering methods like the EM algorithm used in GMM are not applicable for datasets with categorical input attributes \cite {grim2006cluster}}.


\begin{table*}[t]
\caption{Dataset descriptions.}
\label{tbl:datasets}

\resizebox{\textwidth}{!}{%
\begin{tabular}{cccccccccccc}
\hline
\textbf{Type}              & \textbf{Datasets} & \textbf{Attributes} & \textbf{Class 0} & \textbf{Class 1} & \textbf{Class 2} & \textbf{\#Examples} & \textbf{Concept Size} & \textbf{\#Drifts} & \multicolumn{2}{c}{\textbf{Drift Type}} & \textbf{Severity of Each Drift}                          \\ \hline
\multirow{7}{*}{Synthetic} & Virtual 5         & 2                   & 34,3\%           & 35,5\%           & 30,2\%           & 10000               & 2000                  & 5                 & Virtual          & Abrupt               & {[}23.47, 34.1, 24.9, 34.95{]}                           \\ \cline{2-12} 
                           & Virtual 9         & 2                   & 32,6\%           & 35,2\%           & 32,2\%           & 10000               & 1000                  & 9                 & Virtual          & Abrupt               & {[}28.23, 33.9, 28.32, 33.8, 27.23, 32.6, 23.39, 31.6{]} \\ \cline{2-12} 
                           & Circles           & 2                   & 49,8\%           & 50,1\%           & -                & 8000                & 2000                  & 4                 & Virtual/Real     & Incremental          & {[}44.15, 37.55, 32.6{]}                                 \\ \cline{2-12} 
                           & Sine1             & 2                   & 49,8\%           & 50,2\%           & -                & 10000               & 2000                  & 5                 & Real             & Abrupt/Recurrent     & {[}89.7, 88.85, 89.45, 87.75{]}                          \\ \cline{2-12} 
                           & Sine2             & 2                   & 49,5\%           & 50,6\%           & -                & 10000               & 2000                  & 5                 & Real             & Abrupt/Recurrent     & {[}89.8, 88.85, 89.2, 87.75{]}                           \\ \cline{2-12} 
                           & SEA               & 3                   & 50,1\%           & 49,0\%           & -                & 8000                & 2000                  & 4                 & Real             & Gradual              & {[}50.4, 24.15, 47.1{]}                                  \\ \cline{2-12} 
                           & SEAREC            & 3                   & 49,9\%           & 50,0\%           & -                & 16000               & 2000                  & 8                 & Real             & Gradual/Recurrent    & {[}49.4, 24.1, 45.55, 17.95, 47.8, 26.65, 46.2{]}        \\ \hline
\multirow{3}{*}{Real}      & PAKDD             & 29                  & 80,2\%           & 19,7\%           & -                & 50000               & -                     & -                 & -                & -                    & -                                                        \\ \cline{2-12} 
                           & ELEC              & 4                   & 41,5\%           & 58,4\%           & -                & 27549               & -                     & -                 & -                & -                    & -                                                        \\ \cline{2-12} 
                           & NOAA              & 9                   & 31,0\%           & 68,0\%           & -                & 18159               & -                     & -                 & -                & -                    & -                                                        \\ \hline
\end{tabular}
}
\end{table*}

\textbf{Synthetic datasets:} 7 synthetic datasets with various types of drift were used to enable a detailed investigation of how each compared approach behaves in the presence of known drifts. \textcolor{black}{The virtual drift datasets have been proposed in \cite{oliveira2019gmm}. The others were generated using the Tornado framework (Python) proposed in \cite{pesaranghader2018reservoir}\footnote{https://github.com/alipsgh/tornado}. The descriptions of how to generate them are presented in the supplementary material due to space constraints.}

About their characteristics, incremental drifts consist of a steady progression from an old concept to a new one, it can be seen as a sequence of abrupt drifts of low severity. Gradual drifts refers to the transition phase where the probability of observations from the old concept decreases while the probability of the new one increases \cite{minku2009impact}. Severity represents the percentage of the input space which has its target class changed after the drift is complete \cite{minku2009impact}. It was calculated by generating 2000 random instances and checking the percentage of such instances whose labels change from one concept to another, considering 10\% label noise. For virtual drifts, the severity is calculated when a region of the space that previously had $P(\textbf{x}) = 0$ receives observations of a class. Thus it is considered that this region had its target class changed. 



\textbf{Real-world datasets:} Only one modification was made on ELEC dataset, where missing values were removed, resulting in the number of examples in Tbl. \ref{tbl:datasets}. \textcolor{black}{These datasets are also available on-line}\footnote{https://en.wikipedia.org/wiki/Concept\_drift\#Real}. 

\section{Impact of Virtual and Real Drifts on Classifier Suitability}
\label{sec:drift impacts}

This section analyses the impacts of virtual and real drifts on classifiers suitability, answering RQ1. Two representative methods from the literature, IGMM-CD \cite{oliveira2015igmm} and Dynse \cite{almeida2018adapting} (see Section \ref{sec:related work}), are used. Fig. \ref{fig:concept drift impacts} illustrates some representative plots of their execution in the Virtual 9 and Circles datasets \textcolor{black}{before and after a concept drift.} 

\textbf{Virtual drifts} had a different effect on the GMM-based (IGMM-CD) and non-GMM-based (Dynse) classifiers. In particular, we can see from Fig. \ref{fig:dynse virtual9 after} that the decision boundary of Class 2 was incorrectly learned by Dynse, even though a considerable portion of the previously acquired knowledge\footnote{We use the term ``knowledge'' here to refer to the portions of the space considered to belong to each class by the learned classifier.} remained valid after the drift. Therefore, part of the past knowledge acquired for Class 2 must be forgotten in order to modify the decision boundary, while most past knowledge about the Classes 0 and 1 should ideally be retained along with a good portion of the past knowledge of Class 2. Different from Dynse, for IGMM-CD, all past knowledge remained valid after the drift (except for some Gaussians that were incorrectly learned due to noise, which were never valid knowledge), even though such knowledge is insufficient once the virtual drift occurs. This type of model only needs to accommodate additional knowledge when a virtual drift occurs, rather than fixing incorrectly learned knowledge. Accommodating additional knowledge could be done for instance by expanding the decision boundaries of existing Gaussians or adding new Gaussians.


\textbf{Real drifts} always result in a need for forgetting at least part of the previously acquired knowledge. An example of this drift is presented in Fig. \ref{fig:dynse circles after} when the model chosen by Dynse does not reflect the \textit{true} decision boundaries well. DCS-based techniques, according to the selection rule, tend to choose the best classifier from the pool for current data. So, achieving good accuracy depends on good classifiers already trained on the new concept. If such classifiers are unavailable, they will have low accuracy. As this real drift is non-severe, an approach with \textit{on-line} learning like IGMM-CD (Fig. \ref{fig:igmmcd circles after}) was able to keep up faster. However, we can observe a blue Gaussian within the red class, which is a remnant of the old concept learning. Because IGMM-CD does not implement rapid forgetting, old knowledge can cause misclassifications.


Overall, in terms of accuracy drop, we can see that both virtual and real drifts will result in an accuracy drop, given that the \textit{learned} decision boundaries become either incorrect or insufficient. This drop will be smaller/larger depending on the severity of the drift and on how fast the approach can adapt to the drift. However, it is likely that the drop in accuracy for virtual drifts is more in line with the drop in accuracy of non-severe real drifts, as in both cases a good portion of the past knowledge will remain valid. In such cases, adaptation mechanisms able to retain valid past knowledge would enable faster adaptation to the drifts. The drop in accuracy for severe real drifts is likely to be larger, given that a large portion of the previously acquired knowledge will become invalid. In such situation, resetting the system to speed up adaptation to the drift may be useful.

In terms of the need for forgetting past knowledge, for approaches not based on GMM, virtual drifts can have a similar effect to non-severe real drifts as they require forgetting \textit{part} of the past knowledge. Severe real drifts would still have a considerably different effect from virtual drifts on these approaches, as they would require most knowledge to be forgotten, whereas in virtual drifts a good portion of the previously acquired knowledge remains valid. However, for approaches based on GMM, virtual drifts have a considerably different effect from both severe and non-severe drifts, as they do not result in the need for forgetting past knowledge. A good GMM-based approach should be able to benefit from that to adapt to virtual drifts faster.


\section{Proposed Method}
\label{sec:proposed method}

Considering the problems highlighted in Section \ref{sec:drift impacts}, we describe in Sections \ref{subsec:simple model} and \ref{subsec:proposed RQ2}, the components to answer to RQ2, being the adaptation to virtual and real drifts and the noise filter. Finally, in Section \ref{subsec:pool adaptation}, the strategy of pool adaptation to answer RQ3. 

\begin{figure}[t]
	\begin{center}
		\includegraphics[width=3.2in]{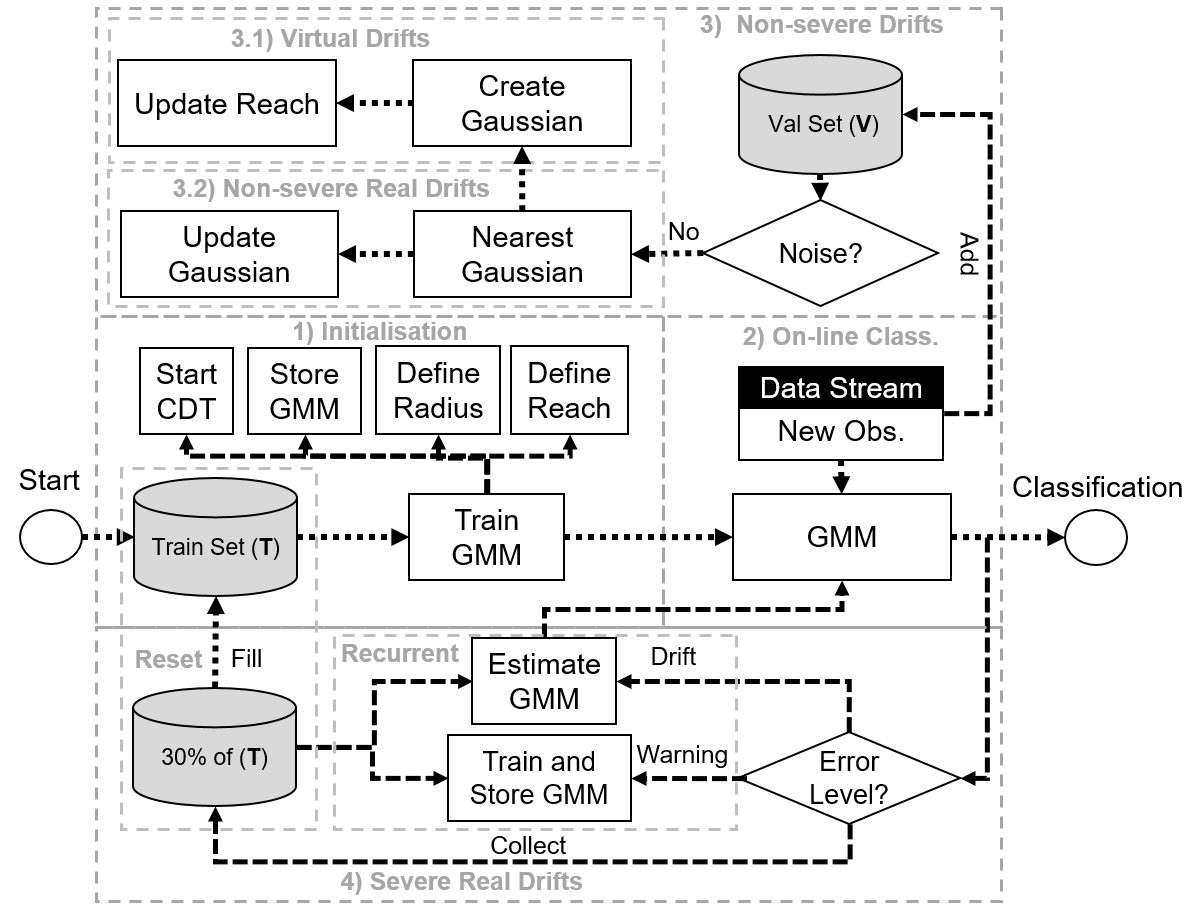}
	\end{center}
	\caption{OGMMF-VRD's overall procedure.}
	\label{fig:proposed method}
\end{figure}

We present in Fig. \ref{fig:proposed method} the general procedure of the OGMMF-VRD. In the first block, the main system mechanisms, such as GMM batch training, are initialized. In the second block, we perform the on-line classification of the observations received from the data stream. In the third block, we update the batch-trained GMM on-line, i.e., using each new observation received. Finally, in the fourth block, we introduce knowledge reset mechanisms for dealing with severe drifts. 

\begin{figure*}[t]
    \centering
    \begin{subfigure}[h]{0.18\textwidth}
        \centering
        \includegraphics[height=1.05in]{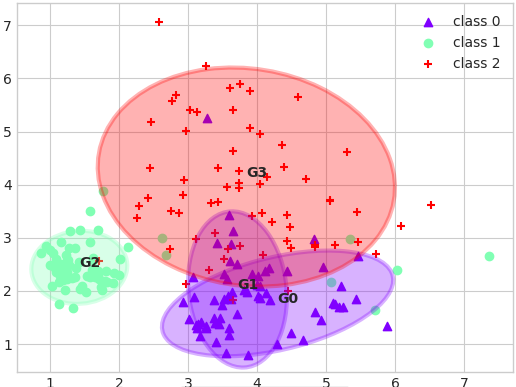}\vspace{-0.2cm}
        \caption{Time: 6050}
        \label{fig:6050}
    \end{subfigure}
    ~ 
    \begin{subfigure}[h]{0.18\textwidth}
        \centering
        \includegraphics[height=1.05in]{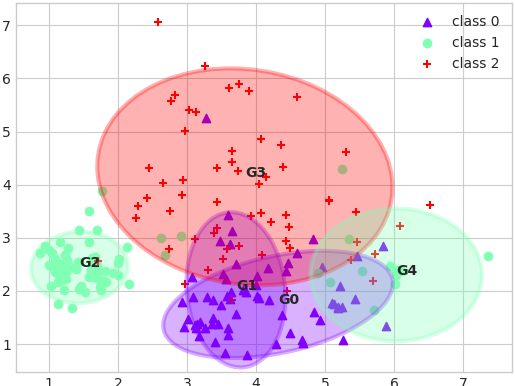}\vspace{-0.2cm}
        \caption{Time: 6080}
        \label{fig:6080}
    \end{subfigure}
    ~ 
    \begin{subfigure}[h]{0.18\textwidth}
        \centering
        \includegraphics[height=1.05in]{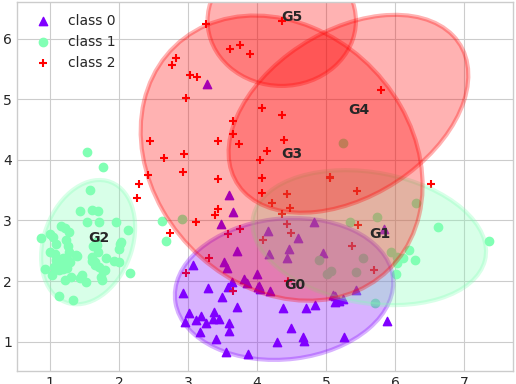}\vspace{-0.2cm}
        \caption{Time: 6110}
        \label{fig:6110}
    \end{subfigure}
    ~ 
    \begin{subfigure}[h]{0.18\textwidth}
        \centering
        \includegraphics[height=1.05in]{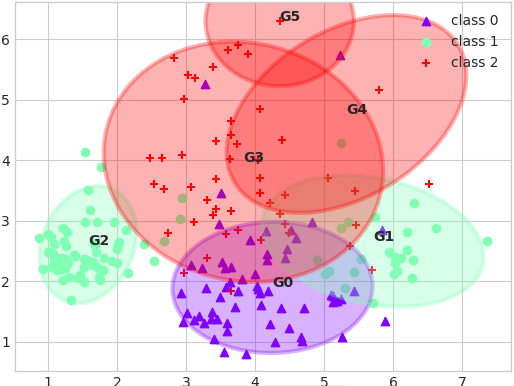}\vspace{-0.2cm}
        \caption{Time: 6140}
        \label{fig:6140}
    \end{subfigure}
    ~ 
    \begin{subfigure}[h]{0.18\textwidth}
        \centering
        \includegraphics[height=1.05in]{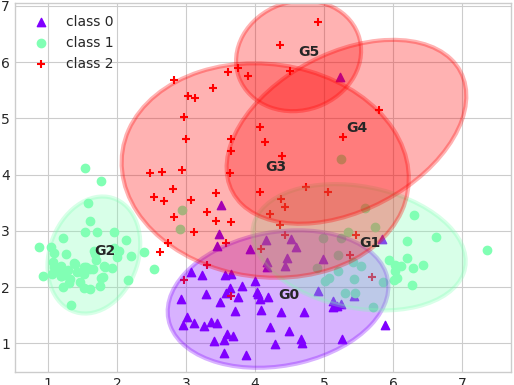}\vspace{-0.2cm}
        \caption{Time: 6170}
        \label{fig:6170}
    \end{subfigure}
    
    \caption{Execution of OGMMF-VRD on dataset Virtual 9. Each image presented illustrates the \textit{learned} decision boundaries over a batch with 200 observations for different time periods. The text G0, ..Gn represent the number of the Gaussian.}
    \label{fig:proposed execution}
\end{figure*}

To clearly understand how OGMMF-VRD works, we present some consecutive plots (Fig. \ref{fig:proposed execution}) of its execution on Virtual 9 dataset, and two videos on youtube, the first being step-by-step\footnote{https://www.youtube.com/watch?v=lP-onPHSR0A} and the second showing plots at every thirty time steps\footnote{https://www.youtube.com/watch?v=D3uZVCFNHyw}.

\subsection{Initialization}
\label{subsec:simple model}



The OGMMF-VRD framework is initialized through two steps: (i) \textbf{GMM training} and (ii) \textbf{drift detector initialization}. 


In \textbf{GMM training}, we collect an initial portion ($m$ observations) of the data stream as the training set ($T$) used to initialize a GMM for each class in $T$ using the algorithm Expectation-Maximization (EM). This algorithm initializes each Gaussian $C_i$ on a random subset from the training set ($T$), and then iteratively adjusts its mean (\bm{$\mu_{i}$}), covariance (\bm{$\Sigma_{i}$}) and weights ($w_{i}$) to maximize the probability of each Gaussian in the distribution modeled. The modeled distribution is given by $P(\textbf{x}) = \sum_{i=1}^{K}P(\textbf{x}|C_{i}) \cdot w_{i}$,
where $K$ is the number of Gaussians and $\textbf{x}$ is a multivariate observation with $d$ dimensions, formally represented by: $\textbf{x}^{d}=\{x_{1}, x_{2},..,x_{d}\}$. Each constant $w_{i}$ is a weight representing the \textcolor{black}{number of observations that constitute the Gaussian $i$, where $0 \leq w_{i} \leq 1$ and $\sum_{i=1}^{K}w_{i} = 1$. $P(\textbf{x}|C_{i})$, represents the conditional probability of observation \bm{$x$} with respect to the Gaussian $C_{i}$}. This probability is computed using the mean (\bm{$\mu_{i}$}) and the covariance (\bm{$\Sigma_{i}$}) of the Gaussian $C_{i}$ as follows:

\begin{equation}
    \label{eq:conditional probability}
    \begin{split}
    P(\textbf{x}|C_{i}) = \frac{1}{(2\pi^{d/2}\sqrt{|\bm{\Sigma_{i}}|})}exp(-\frac{1}{2}(\bm{x}-\bm{\mu_{i}})^{T}\bm{\Sigma_{i}^{-1}}(\bm{x}-\bm{\mu_{i}}))
    \end{split}
\end{equation}


In order to define the optimum number of Gaussians, for each class, we train different GMMs ranging the number of Gaussians from 1 to $Kmax$. The best resulting GMM model is chosen using the higher value of AIC criterion and is added to the final GMM model. The AIC criterion is defined by $2 \cdot p - 2 \cdot L$. Here, $p$ represents the number of model parameters. In a GMM with only one Gaussian, the parameters are the mean (\bm{$\mu_{i}$}), covariance (\bm{$\Sigma_{i}$}), and weight ($\omega_{i}$). So, for each existing Gaussian in a GMM, the value of $p$ is multiplied by three. The parameter $L$ represents the maximum likelihood function of the GMM on a set of $m$ observations, defined by Eq. \ref{eq:L}:

\begin{equation}
    \label{eq:L}
    L = \sum_{i=1}^{m}log\sum_{j=1}^{K}P(\bm{x_{i}}|C_{j}) \cdot w_{j}
\end{equation}

The final GMM will be used to predict the labels of the incoming observations (\bm{$x_{t}$}) in Part 2 (\textit{On-line} classification) of Fig. \ref{fig:proposed method}. The predicted label \textcolor{black}{can be used by users for decision-making, using $\hat{y} = \argmax_{i \in \{1,2,\cdots,K\}}P(C_{i}|\bm{x})$}, 
where $\hat{y}$ represents the predicted label and $P(C_{i}|\bm{x})$ represents the posterior probability of a Gaussian $C_{i}$ given the observation \bm{$x$}, as defined by Eq. \ref{eq:posterior probability}. Thus, the incoming observation (\bm{$x_{t}$}) is classified with a class of the Gaussian that presented for it the higher posterior probability.

\begin{equation}
    \label{eq:posterior probability}
    P(C_{i}|\bm{x}) = \frac{P(\bm{x}|C_{i})\cdot w_{i}}{\sum_{i=1}^{K}P(\bm{x}|C_{i})\cdot w_{i}}
\end{equation}

In \textbf{drift detector initialization}, we start a CDT to monitor the system error and identify performance degradation. In an experiment we evaluate ECDD \cite{ross2012exponentially}, CUSUM \cite{page1954continuous}, FHDDM \cite{pesaranghader2016fast}, DDM \cite{gama2004learning}, and EDDM \cite{baena2006early}. EDDM reached the best results and so was chosen. After initialization, non-severe drifts are treated as in Section \ref{subsubsec:virtual adaptation} and severe drifts as in Section \ref{subsubsec:real adaptation}.

\subsection{Coping with Virtual and Real Drifts While Achieving Robustness to Noise}
\label{subsec:proposed RQ2}

\subsubsection{Dealing with Virtual Drifts and Non-Severe Real Drifts}
\label{subsubsec:virtual adaptation}

This part of the OGMMF-VRD aims to maintain the useful knowledge of the system in the presence of virtual and non-severe real drifts. Alg. \ref{alg:virtual adaptation} presents the overall procedure.

\begin{algorithm}[H]
	\centering
	\caption{NonSevereDriftAdaptation()}
	\label{alg:virtual adaptation}	
	\begin{algorithmic}[1]
		\Statex \textbf{Input:} observation (\bm{$x_{t}$}, $y_{t}$)
		\State gaussian, pertinence $\leftarrow$ GaussianClose(\bm{$x_{t}$}, $y_{t}$)
		\State UpdateGaussian(gaussian, \bm{$x_{t}$})
    	\If{pertinence $> \theta$}
    		\State CreateGaussian(\bm{$x_{t}$}, $y_{t}$)
    		\State UpdateReach(pertinence)
    	\EndIf
	\end{algorithmic}
\end{algorithm}

In Line 1, Gaussian Close represents the mechanism used to determine when to create a new Gaussian, and which Gaussian will be updated. For each incoming observation (\bm{$x_{t}$}), we need to know where it is located in relation to existing Gaussians. For that, we use Alg. \ref{alg:gaussian near}.

\begin{algorithm}[H]
	\centering
	\caption{GaussianClose()}
	\label{alg:gaussian near}	
	\begin{algorithmic}[1]
		\Statex \textbf{Input:} observation (\bm{$x_{t}$}, $y_{t}$)
		\Statex \textbf{Output:} gaussian, pertinence
			\State aux $\leftarrow$ $\varnothing$
			\For {each $gaussian$ in GMM}
				\If{$gaussian$ $\in$ $y_{t}$}
					\State aux.append($P(\bm{x_{t}}|gaussian)$) \Comment Eq. \ref{eq:conditional probability}
				\Else
					\State aux.append(0)
				\EndIf
			\EndFor
			\State ngaussian $\leftarrow$ argmax(aux) \Comment{Nearest gaussian}
			\State pertinence $\leftarrow$ aux[$ngaussian$] \Comment{Pertinence of $x_{t}$}
		\end{algorithmic}
	\end{algorithm}

This routine computes the conditional probability (Eq. \ref{eq:conditional probability}) of an incoming observation (\bm{$x_{t}$}) for all existing Gaussians with the same class as the incoming observation ($y_{t}$) (Lines 1 to 8). This algorithm returns the nearest Gaussian with the same class of the incoming observation (Line 9), and the pertinence of the incoming observation to the Gaussians (Line 10). With this information, we can trigger appropriate drift handling strategies (Sections \ref{subsubsubsec:real drifts adaptation} and \ref{subsubsubsec:virtual drifts adaptation}). 

\paragraph{\textbf{Adjusting Existing Gaussians}}
\label{subsubsubsec:real drifts adaptation}

Line 2 (Gaussian Update) of Alg. \ref{alg:virtual adaptation}, represents the process to adapt to non-severe real drifts. The idea is that non-severe real drifts change little the decision boundaries of the problem, so a simple displacement of existing Gaussians can handle this. 
Therefore, we used the modifications to the equations of the EM algorithm proposed by Engel et al.~\cite{engel2010incremental}. These modifications can update a Gaussian based on a single new observation (\bm{$x_{t}$}) and its current parameters (mean \bm{$\mu_{i}^{t}$}, covariance \bm{$\Sigma_{i}^{t}$} and weight $w_{i}^{t}$). Another parameter necessary is the variable $sp_{i}^{t}$, \textcolor{black}{defined by $sp_{i}^{t} = sp_{i}^{t-1} + P(C_{i}|\bm{x})$}, that will store the accumulated posterior probability (Eq. \ref{eq:posterior probability}) of each Gaussian. Thus, the Equations used to update the Gaussian parameters are shown below \cite{engel2010incremental}: 


\begin{equation}
    \label{eq:weight incremental}
    w_{i}^{t} = \frac{sp_{i}^{t}}{\sum_{j}^{K}sp_{j}^{t}} 
\end{equation}

\begin{equation}
    \label{eq:mean incremental}
    \bm{\mu_{i}^{t}} = \bm{\mu_{i}^{t-1}} + \frac{P(C_{i}|\bm{x})}{sp_{i}^{t}} \cdot (\bm{x} - \bm{\mu_{i}^{t-1}})
\end{equation}

\begin{equation}
    \label{eq:covariance incremental}
    \begin{split}
    \bm{\Sigma_{i}^{t}} = \bm{\Sigma_{i}^{t-1}} - (\bm{\mu_{i}^{t}} - \bm{\mu_{i}^{t-1}})^{T}(\bm{\mu_{i}^{t}} - \bm{\mu_{i}^{t-1}}) \\+ \frac{P(C_{i}|\bm{x})}{sp_{i}^{t}} \cdot [\bm{\Sigma_{i}^{t-1}} - (\bm{x} - \bm{\mu_{i}^{t}})^{T}(\bm{x} - \bm{\mu_{i}^{t}})]
    \end{split}
\end{equation}

\paragraph{\textbf{Adding New Gaussians}}
\label{subsubsubsec:virtual drifts adaptation}

Line 4 (Create Gaussian) of Alg. \ref{alg:virtual adaptation} represents the process to adapt to virtual drifts. Since virtual drifts do not change the true decision boundaries of the problem, new Gaussians can be used to accommodate new data that is far from existing Gaussians. For this, when the pertinence of the incoming observation (\bm{$x_{t}$}) is less than $\theta$ (line 3 of Alg. \ref{alg:virtual adaptation}), we consider that this observation is far away from existing Gaussians and it is necessary to create another Gaussian to represent the new region in the feature space. Theta ($\theta$) is the lowest pertinence (Eq. \ref{eq:conditional probability}) obtained from the observations in the training set ($T$). Points more distant than theta ($\theta$) are out of GMM's reach. 




Thus, the new Gaussian is initialized using $sp_{i}$ = $w_{i} = 1$, $\bm{\mu_{i}} = \bm{x_{i}}$ and $\bm{\Sigma_{i}} = \textit{Cfc} \cdot \bm{I}$, where \bm{$I$} represents the identity matrix, which has the same number of dimensions as \bm{$x_{t}$}, and $\textit{Cfc}$ represents the size of the Gaussian circumference \textcolor{black}{defined by 
$\textit{Cfc} = (x_{max} - x_{min})/20$}, 
where $x_{max}$ and $x_{min}$ represent the highest and lowest observed value for attributes in the initialization training set ($T$), and 20 was fixed for all datasets, but can be adjusted (block \textbf{Define Radius} in Fig. \ref{fig:proposed method}). After the initialization of the new Gaussian's parameters, we re-normalize the weights of all existing Gaussians using Eq. \ref{eq:weight incremental}. 

Line 5 (Update Reach) represents the process of updating $\theta$. For that, we use the pertinence of the incoming observation to substitute the older value (Line 1 of Alg \ref{alg:virtual adaptation}). This update indicates that, for the creation of new Gaussians, the incoming observation must have a lower pertinence than theta ($\theta$), indicating that it is farther away.


\subsubsection{Dealing With Severe Real Drifts}
\label{subsubsec:real adaptation}

This part of the OGMMF-VRD aims to reset useless system knowledge. Given the adoption of the mechanisms explained in \ref{subsubsec:virtual adaptation}, non-severe drifts will not degrade the system's predictive performance so much as severe real drifts. Therefore, we consider that severe real drifts will significantly degrade the system's performance, and such degradation can be used to detect such drifts. \textcolor{black}{Thus, to identify these degradations, we will use a CDT, which monitors the classification performance of the system using the error obtained for each new observation. If the system error rises above a threshold, the CDT reports a concept drift, indicating that system knowledge is obsolete. At that time new observations of the data streams are stored and used to retrain the whole system.}

OGMMF-VRD thus detects drifts by using the Early Drift Detection Method (EDDM) \cite{baena2006early}. If EDDM detects a drift, the system collects new data to learn its knowledge from scratch. EDDM has as parameters the tolerance levels defined by warning ($w$) and drift ($c$) levels, and can be tuned so that it only detects severe drifts, which will degrade the system's performance more than non-severe drifts. Each tolerance level yields a different response: (i) for the normal error level the model remains untouched; (ii) for the warning error level, the system begins to collect incoming observations to retrain the model; and (iii) for the drift level, new observations are collected and added to the observations collected during the warning level to fill a batch with $m$ new observations to compose a new training set $T$ to initialize a new model.

\subsubsection{Noise Filter}
\label{subsec:noise filter}

In this Section, we will discuss how to address the part of RQ2 which seeks to know how to achieve robustness to noise in virtual and real drifts. Noisy observations cause two main problems to GMM models: (i) if an observation is too far from its class, the system tends to create a new Gaussian. If this observation was noisy, it means that a Gaussian of an undesired class will be created in the region belonging to another class; (ii) if a Gaussian is updated on a noisy observation, it will move to a region of the space that does not correspond to its true boundary. These points may impair the performance of the GMM over time. To overcome them, we use  k-Disagreeing Neighbors (kDN) \cite{walmsley2018ensemble}, defined in Eq. \ref{eq:kdn}, as a noise filter.  
 

\begin{equation}
    \label{eq:kdn}
   \textcolor{black}{kDN(\bm{x}) = \frac{|\bm{\forall \bm{x'} | \bm{x'} \in kNN(\bm{x}) \wedge label(\bm{x')}}|}{k}}
\end{equation}


Consider that we need to determine if an observation $(\bm{x},y)$ is noise. The kDN represents the fraction of the k nearest neighbors of this observation that do not share the same class $y$. Values close to 0 indicate that $(\bm{x},y)$ is easy to classify and unlikely to be noise, and close to 1 indicate that it $(\bm{x},y)$ is difficult to classify and may be noise. 

%

In OGMMF-VRD, we use the noise filter in two parts: (i) during the GMM initialization and (ii) before the non-severe drift adaptation (Section \ref{subsubsec:virtual adaptation}). In the GMM initialization, we use kDN as a \textit{pre-processing} to remove all noisy observations of the initialization training set ($T$). 


Before the non-severe drift adaptation (Block Noise on Part 3 of Fig. \ref{fig:proposed method}), we use  kDN to avoid updates using noisy observations. To do this \textit{on-line}, we use a validation set ($V$). This set has the same size as the training set ($T$) used to initialize the GMMs. The difference between them is that the $T$ set is only used for initializing the GMMs (Part 1 of Fig. \ref{fig:proposed method}) and the $V$ set acts like a sliding window (Part 3 of Fig. \ref{fig:proposed method}), on which for each new observation inserted an older one is removed. 

Note that the validation set ($V$) helps to distinguish between noise-free observations, noisy observations and gradual drifts. A noise-free observation has many neighbors with the same class. A noisy observation is an observation that appears in a region with a different class from its own class only once. Gradual drifts are a set of observations that over time are more likely to appear in a new region. Since we store all data in the validation set ($V$), we will be able to verify the growth of observations in a new region, thus avoiding confusion between gradual drift and noise.

For our approach, we have specified the neighborhood to $k=5$, and observations with kDN greater than 0.8 should be avoided because they have 80\% of their neighborhood with a different class which can strongly indicate that it is a noise, and can hinder the generalization of the system.

\subsection{Harnessing Knowledge From Past Similar Concepts by Using a Pool Adaptation Strategy}
\label{subsec:pool adaptation}

This section presents our method proposed to answer RQ3, which seeks to harness past knowledge to accelerate adaptation to both virtual and real drifts. For this, we use a pool ($P$) to store previously trained GMMs. GMMs are re-initialized at two occasions: (i) when EDDM \cite{baena2006early} reaches drift level (c) indicating that the system must be reset, i.e. after collecting $m$ observations from the data stream to use as a new training set ($T$); and (ii) when EDDM reaches warning level (w) indicating that a severe drift may be occurring (see Section \ref{subsubsec:real adaptation}), i.e. when 30\% of $m$ are collected. This is represented by block \textbf{Store GMM} in Fig. \ref{fig:proposed method}.

Classifiers are estimated from the pool when EDDM \cite{baena2006early} reaches the drift level (c). So to avoid waiting for all $m$ observations used for retraining, when we get 30\% of $m$, we select the classifier from the pool with the best accuracy for this data and use it to replace the current model. This is represented by block \textbf{Estimate GMM} in Fig. \ref{fig:proposed method}.

If the  maximum number of models in $P$ is reached, the oldest GMM is removed from $P$. Some preliminary experiments were done and the maximum pool size of 20 obtained the best cost-benefit between accuracy and runtime.

\section{Experiments and Discussions}
\label{sec:experiments}

Three experiments were realized in this work: (i) comparison with literature works (Section \ref{subsec:comparison}); (ii) noise filter robustness (Section \ref{subsec:noise analysis}); and (iii) proposed method mechanisms analysis (Section \ref{subsec:mechanism analysis}). All experiments were evaluated using the datasets discussed in Section \ref{subsec:datasets} and the metrics discussed in Section \ref{subsec:exp metrics}.

\subsection{Metrics and Statistical Tests}
\label{subsec:exp metrics}

To evaluate the predictive performance of the approaches, we used the following metrics: 

\textbf{Cross Validation for Data Streams:} traditional cross-validation cannot be applied to data streams because it splits data randomly, making it impossible to see concepts in the order that they should be viewed. Therefore, we use a modified version proposed in \cite{sun2016online}. It consists of leaving out an observation every period X to construct the stream. For example, we remove the first element within the first thirty, then remove the first element from the thirty second and so on. For all datasets, 30 runs were executed, i.e. X=30. The element removed corresponds to the order of the execution.

\textbf{Overall Accuracy and G-mean:} both are calculated based on the on-line predictions given by the system. Accuracy has been used in several data stream learning studies, such as \cite{sun2016online, oliveira2015igmm, almeida2018adapting}. G-mean is the geometric mean of the recall on each class, and is a metric independent of the level of class imbalance in the data \cite{wang2018systematic}.  


%




\textbf{Accuracy Over Time (AOT):} this is the time series showing the system's accuracy over time \cite{sun2016online, oliveira2015igmm, almeida2018adapting}, where each value represents the accuracy over a batch. 
For example, if the batch size is X = 250: $batch_{1}$ = [0 to 250), $batch_{2}$ = [251 to 500), $batch_{3}$ = [501 to 750), etc, until the end of the data stream. In order to increase the discriminative power of the metric, the standard deviation between the several accuracies is also reported.



\textbf{Runtime:} the execution time of the approaches was measured in seconds considering the difference between the end time and the start time on a machine with 8GB of ram and processor Intel Xeon E3-1220 v5.

\textbf{Friedman and Nemenyi Tests:}  Friedman is a non-parametric statistical hypothesis test that can be used to compare multiple approaches across multiple datasets \cite{demvsar2006statistical}. It ranks the algorithms and checks whether the null hypothesis that they are all equal can be rejected. If the null hypothesis is rejected, then the Nemenyi post-hoc is used to check which of the approaches is significantly different from each other. Both tests were used in this work for a significance level of $\alpha$ = 0.05.

\textbf{Wilcoxon Test:} this is a non-parametric statistical hypothesis test used in our experiments to evaluate how a specific mechanism has improved over the accuracy of the complete system. 
This test was used with a significance level of $\alpha$ = 0.05.


\subsection{Comparison With Existing Approaches}
\label{subsec:comparison}

This experiment aims to validate the performance of the OGMMF-VRD in comparison with literature works. The discussions are divided in Section \ref{subsubsec:approaches virtual and real}, literature approaches that claim to deal with both virtual and real drifts, and Section \ref{subsubsec:approaches concept drift}, literature approaches that deal with concept drift.

For the first comparison, we selected: 
IGMM-CD \cite{oliveira2015igmm}, Dynse \cite{almeida2018adapting} and GMM-VRD \cite{oliveira2019gmm} (see Section \ref{sec:related work}). The parameters used for the approaches are shown in 
Tbl. \ref{tbl:parameters}. \textcolor{black}{An analysis of sensitivity of OGMMF-VRD to the radius (Gaussian Circumference) and c (drift level for EDDM) parameters is presented in the supplementary material, due to space constraints, and shows that these parameters do not significantly affect OGMMF-VRD's accuracy.}

\textcolor{black}{For the second comparison, we  selected two groups of approaches (i) ensemble methods and (ii) drift detectors with an incremental algorithm. Methods based on ensembles are: Accuracy Weighted Ensemble (AWE) \cite{wang2003mining}, Adaptive Random Forest (ARF) \cite{gomes2017adaptive}, Leveraging Bagging ensemble classifier (LevBag) \cite{bifet2010leveraging}, and Oza Bagging Ensemble classifier (OzaAS) with and without ADWIN drift detector (OzaAD) \cite{oza2005online}. Methods based on drift detectors are: ADWIN \cite{bifet2007learning}, DDM \cite{gama2004learning}, and EDDM \cite{baena2006early}, combined with Hoeffding Adaptive Tree classifier (HAT) \cite{bifet2009adaptive}. All of these approaches are available on-line in the library scikit-multiflow\footnote{https://scikit-multiflow.readthedocs.io/en/stable/index.html} and the parameters used in the comparison were the default values provided by the library.}

\begin{table}[t]
\caption{Parameters used for the compared approaches. A grid search was executed for the most important parameters and the best was chosen considering the average accuracy across datasets.}
\label{tbl:parameters}

\resizebox{\columnwidth}{!}{%
\begin{tabular}{ccccc}
\hline
\textbf{Algorithm}         & \textbf{Parameters} & \textbf{Grid Search}         & \textbf{Synthetic}   & \textbf{Real} \\ \hline
\multirow{3}{*}{IGMM-CD}   & Sigma\_ini          & {[}0.5, 1, 2, 5, 10{]}       & 0.05                 & 10            \\ \cline{2-5} 
                           & Cver                & -                            & 0.01                 & =             \\ \cline{2-5} 
                           & T                   & {[}1, 5, 7, 9, 13{]}         & 13                   & =             \\ \hline
\multirow{7}{*}{Dynse}     & D                   & -                            & 25                   & =             \\ \cline{2-5} 
                           & m                   & {[}50, 100, 200, 300, 400{]} & 50                  & =             \\ \cline{2-5} 
                           & M                   & -                            & 100                   & =             \\ \cline{2-5} 
                           & k                   & -                            & 5                    & =             \\ \cline{2-5} 
                           & CE                  & -                            & A Priori             & =             \\ \cline{2-5} 
                           & PE                  & -                            & Age Based            & =             \\ \cline{2-5} 
                           & BC                  & -                            & Gaussian Naive Bayes & =             \\ \hline
\multirow{7}{*}{GMM-VRD}   & m                   & {[}50, 100, 200, 300, 400{]} & 50                   & 200           \\ \cline{2-5} 
                           & EM it.              & -                            & 10                   & =             \\ \cline{2-5} 
                           & kmax                & {[}2, 4, 6, 8{]}             & 2                    & =             \\ \cline{2-5} 
                           & kDN                 & -                            & 5                    & =             \\ \cline{2-5} 
                           & Detector            & -                            & ECDD                 & =             \\ \cline{2-5} 
                           & c                   & -                            & 1                    & =             \\ \cline{2-5} 
                           & w                   & -                            & 0.5                  & =             \\ \hline
\multirow{3}{*}{OGMMF-VRD} & P                   & -                            & 20                   & =             \\ \cline{2-5} 
                           & radius              & {[}10, 15, 20, 25{]}         & 20                   & =             \\ \cline{2-5} 
                           & c                   & {[}1, 1.5, 2, 2.5{]}         & 1                    & =             \\ \hline
\end{tabular}
}
\end{table}

\begin{figure*}[t]
    \centering
    \begin{subfigure}[h]{0.35\textwidth}
        \centering
        \includegraphics[height=0.72in]{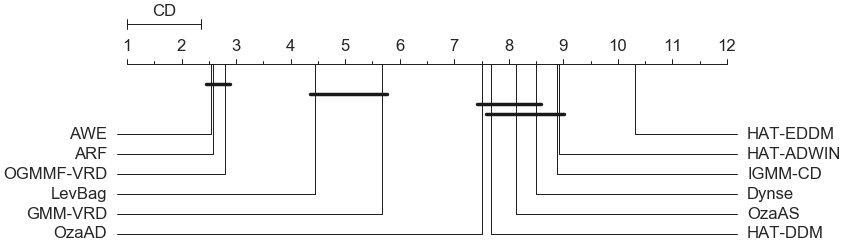}
        \caption{\textcolor{black}{Accuracy}}
        \label{fig:friedman accuracy}
    \end{subfigure}
    ~ 
    \begin{subfigure}[h]{0.3\textwidth}
        \centering
        \includegraphics[height=0.72in]{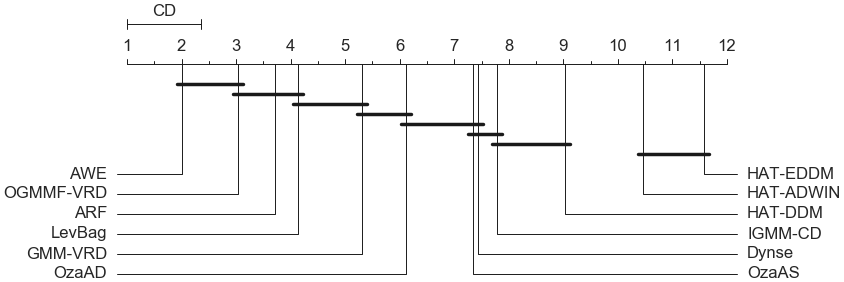}
        \caption{\textcolor{black}{G-mean}}
        \label{fig:friedman gmean}
    \end{subfigure}
    ~ 
    \begin{subfigure}[h]{0.32\textwidth}
        \centering
        \includegraphics[height=0.82in]{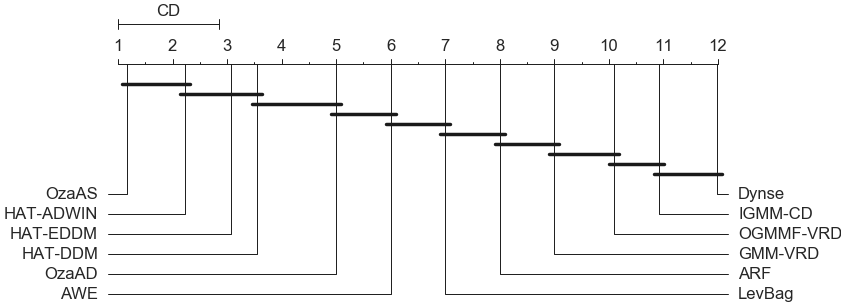}
        \caption{\textcolor{black}{Runtime}}
        \label{fig:friedman time}
    \end{subfigure}
    
    \caption{Friedman ranking from left to right. Friedman's p-values were 1.96E-181, 1.04E-205 and 3.06E-239, respectively, indicating rejection of the null hypothesis at the level of significance of $\alpha=0.05$. Any pair of approaches whose distance between them is larger than CD is considered to be different according to the Nemenyi posthoc tests.}
  \label{fig:friedman}
\end{figure*}



\subsubsection{Approaches that Handle Virtual and Real Drifts}
\label{subsubsec:approaches virtual and real}

\begin{figure}[t]
    \centering
    \begin{subfigure}[h]{0.5\textwidth}
        \centering
        \includegraphics[height=2.4in]{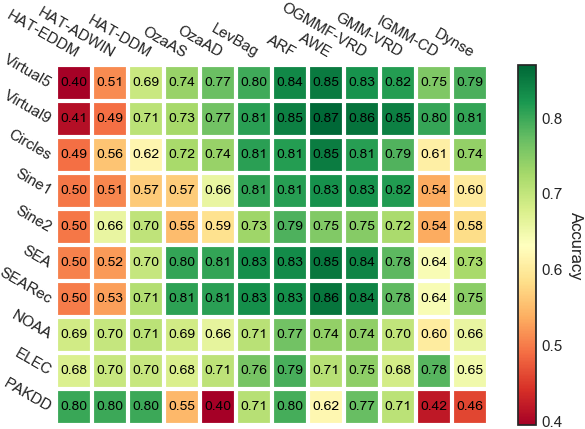}
        \caption{\textcolor{black}{Accuracy}}
        \label{fig:average accuracy}
    \end{subfigure}
    ~ 
    \begin{subfigure}[h]{0.5\textwidth}
        \centering
        \includegraphics[height=2.4in]{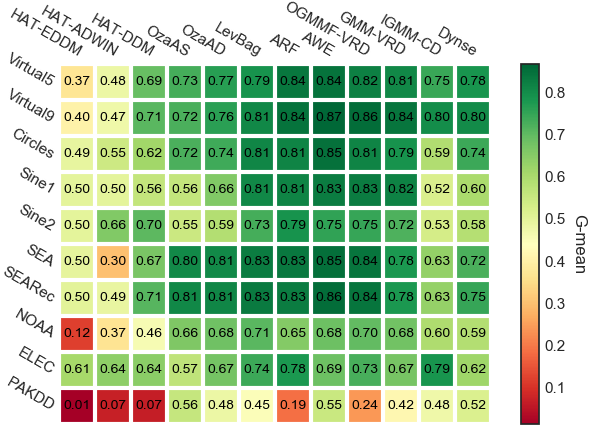}
        \caption{\textcolor{black}{G-mean}}
        \label{fig:average gmean}
    \end{subfigure}
    ~ 
    \begin{subfigure}[h]{0.5\textwidth}
        \centering
        \includegraphics[height=2.4in]{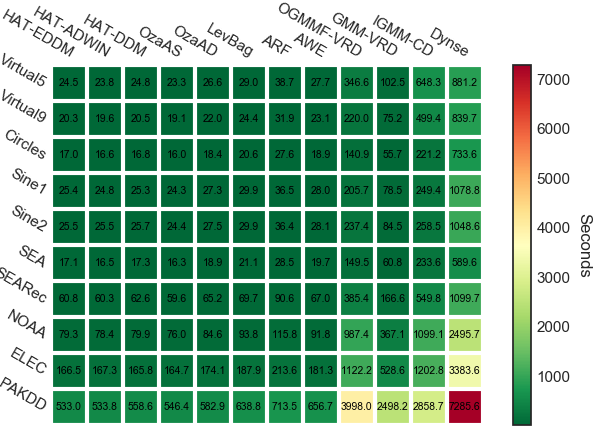}
        \caption{\textcolor{black}{Runtime}}
        \label{fig:average time}
    \end{subfigure}
    
    \caption{Overall Performance Average. Cells closer to green/red represent better/worse results.}
    \label{fig:average results}
\end{figure}

A heat map is shown in Fig. \ref{fig:average results} for the compared approaches using both synthetic and real-world datasets.  To attest the statistical difference of the results, we present in Fig. \ref{fig:friedman} the rank of Friedman with the Nemenyi post-hoc for all metrics evaluated. 
For all performance metrics, Friedman rejected the null hypothesis that the algorithms have equal performance at the level of significance of $\alpha=0.05$. 
According to the Nemenyi post-hoc tests for average accuracy (Fig. \ref{fig:friedman accuracy}), \textcolor{black}{OGMMF-VRD is statistically significantly different from GMM-VRD, IGMM-CD and Dynse.} For average G-mean (Fig. \ref{fig:friedman gmean}), only Dynse and IGMM-CD were not found to be statistically different from each other. For runtime (Fig. \ref{fig:friedman time}), only OGMMF-VRD and IGMM-CD.  To provide a more detailed understanding of these results, Fig. \ref{fig:accuracy over time} \textcolor{black}{presents in the first row of figures some plots of AOT for GMM-VRD, IGMM-CD, and Dynse.  AOTs for the rest of the datasets are presented in the supplementary material due to space constraints.}

Looking at the dataset which has abrupt virtual drifts (Virtual 9  in Fig. \ref{fig:virtual9 accuracy vr}), we can see that IGMM-CD and GMM-VRD had good AOT. This is because these datasets have similarities between their concepts, and despite this approach potentially generating too many Gaussians over time can handle this type of drift properly. Regarding Dynse, its AOT has high peaks indicating that the choice of good classifiers from the pool improves the results. However, in the presence of drifts, its performance declines. OGMMF-VRD presents high accuracy almost all the time. One reason for this is its ability to quickly find out which new regions of space need to be learned.

We now look at the datasets with real drifts: Circles (Fig. \ref{fig:circles accuracy vr}), which has virtual/real drifts, and Sine 2 (Fig. \ref{fig:sine2 accuracy vr}), which has abrupt/recurrent shifts, we see that Dynse has the biggest drop in accuracy compared to other approaches. This is explained by two points: (i) in the presence of a new concept, Dynse's accuracy only rises when several classifiers are trained on the new concept; (ii) if its pool is small, over time older classifiers are deleted when a new one is added, so if a similar concept is slow to appear the pool may no longer have a suitable model. Regarding IGMM-CD, it is observed that its accuracy drops dramatically and hardly goes back up. This is because IGMM-CD does not have a fast reset mechanism, and it forces the system to spend a lot of time with obsolete Gaussians. Regarding GMM-VRD and OGMMF-VRD, it is observed that in the presence of drifts their accuracy does not decrease so much in relation to other approaches. This is because it has a CDT that informs quickly when their performance is deteriorating, allowing it to be reset to fit the new concept. 

Now, we look at the real-world datasets: NOAA (Fig. \ref{fig:noaa accuracy vr}), and PAKDD (Fig. \ref{fig:pakdd accuracy vr}).  For the NOAA, it is observed that Dynse and IGMM-CD sometimes decline their accuracy, which demonstrates that these methods may not be robust to different types of drift. For PAKDD, GMM-VRD and OGMMF-VRD obtained very good AOT, away from that of the other approaches. One reason for this is that both of these methods have a model selection mechanism targeted at improving accuracy in the initialization phase. In addition, they implement a fast reset whenever necessary, helping them to  always maintain good accuracy. As other approaches do not have these mechanisms, they tend to have poor AOT.

\subsubsection{Approaches that Deal with Concept Drift}
\label{subsubsec:approaches concept drift}

\begin{figure*}[t]
    \centering
     \begin{subfigure}[h]{0.18\textwidth}
        \centering
        \includegraphics[height=1.05in]{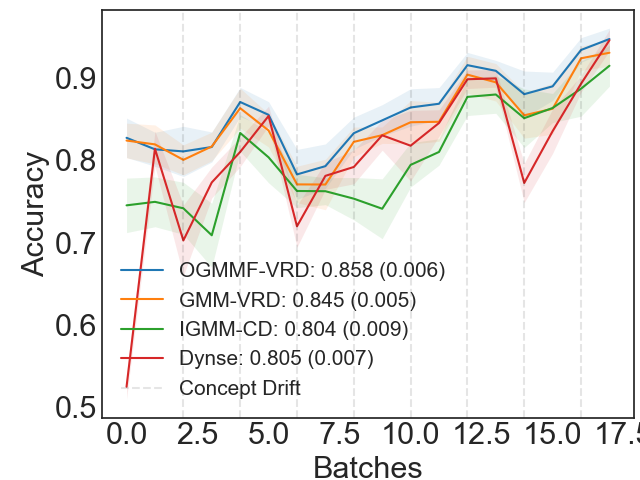}\vspace{-0.2cm}
        \caption{Virtual 9 (VR)}
        \label{fig:virtual9 accuracy vr}
    \end{subfigure}
    ~ 
    \begin{subfigure}[h]{0.18\textwidth}
        \centering
        \includegraphics[height=1.05in]{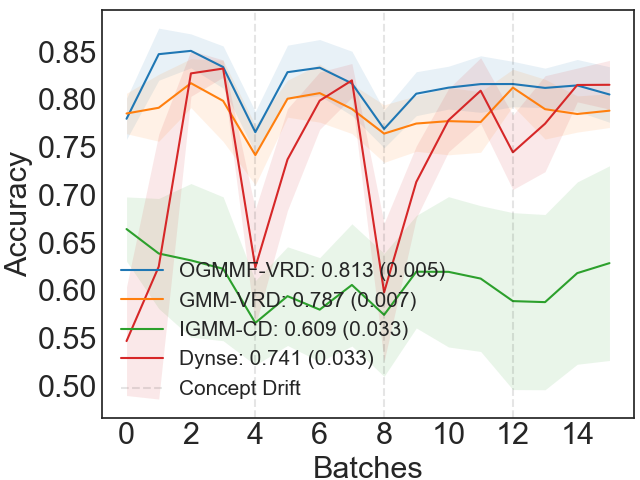}\vspace{-0.2cm}
        \caption{Circles (VR)}
        \label{fig:circles accuracy vr}
    \end{subfigure}
    ~
    \begin{subfigure}[h]{0.18\textwidth}
        \centering
        \includegraphics[height=1.05in]{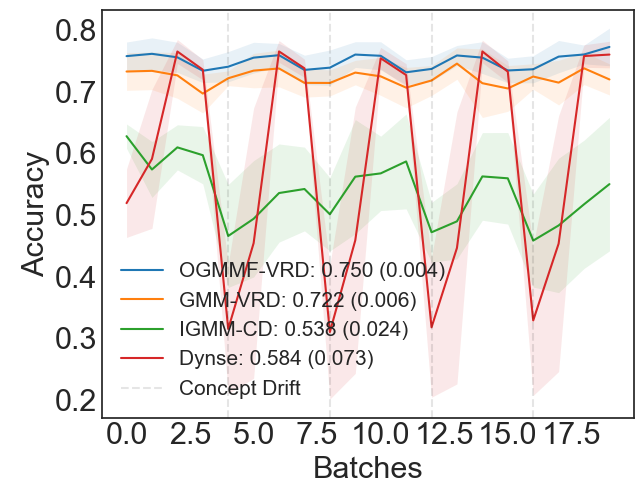}\vspace{-0.2cm}
        \caption{Sine 2 (VR)}
        \label{fig:sine2 accuracy vr}
    \end{subfigure}
    ~ 
    \begin{subfigure}[h]{0.18\textwidth}
        \centering
        \includegraphics[height=1.05in]{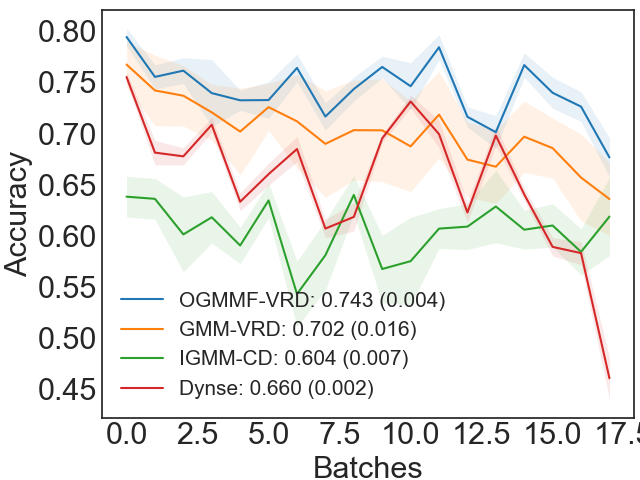}\vspace{-0.2cm}
        \caption{NOAA (VR)}
        \label{fig:noaa accuracy vr}
    \end{subfigure}
    ~ 
    \begin{subfigure}[h]{0.18\textwidth}
        \centering
        \includegraphics[height=1.05in]{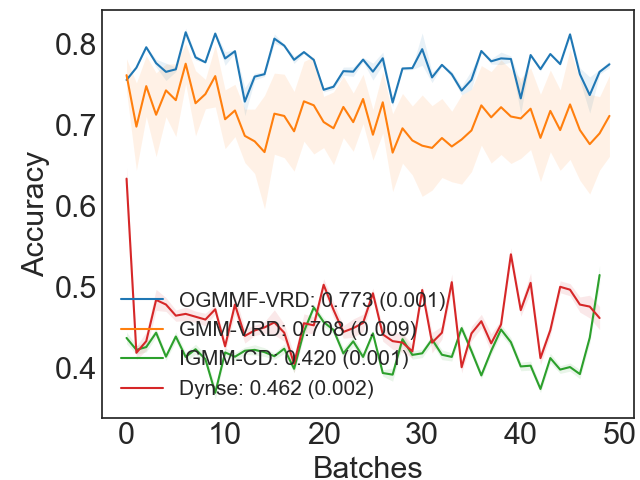}\vspace{-0.2cm}
        \caption{PAKDD (VR)}
        \label{fig:pakdd accuracy vr}
    \end{subfigure}
    ~ 
    \begin{subfigure}[h]{0.18\textwidth}
        \centering
        \includegraphics[height=1.05in]{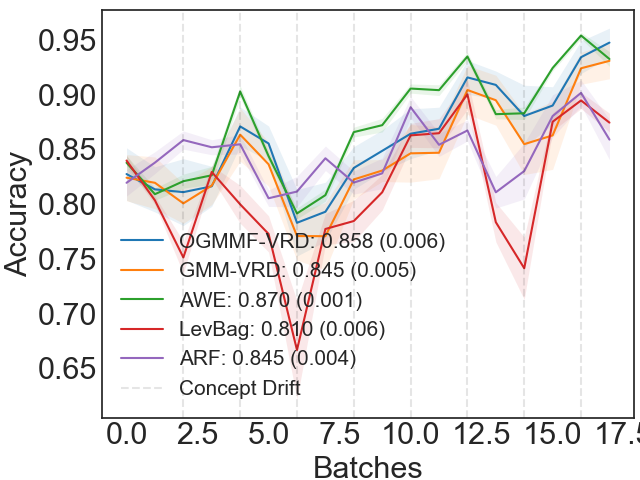}\vspace{-0.2cm}
        \caption{Virtual 9 (Bests)}
        \label{fig:virtual9 accuracy bests}
    \end{subfigure}
    ~ 
    \begin{subfigure}[h]{0.18\textwidth}
        \centering
        \includegraphics[height=1.05in]{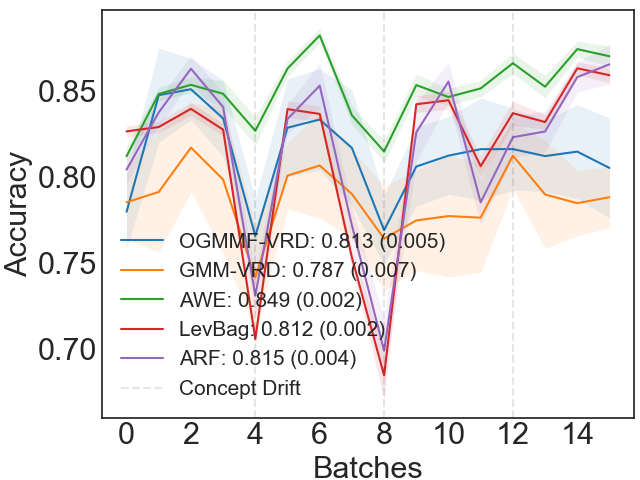}\vspace{-0.2cm}
        \caption{Circles (Bests)}
        \label{fig:circles accuracy bests}
    \end{subfigure}
    ~ 
    \begin{subfigure}[h]{0.18\textwidth}
        \centering
        \includegraphics[height=1.05in]{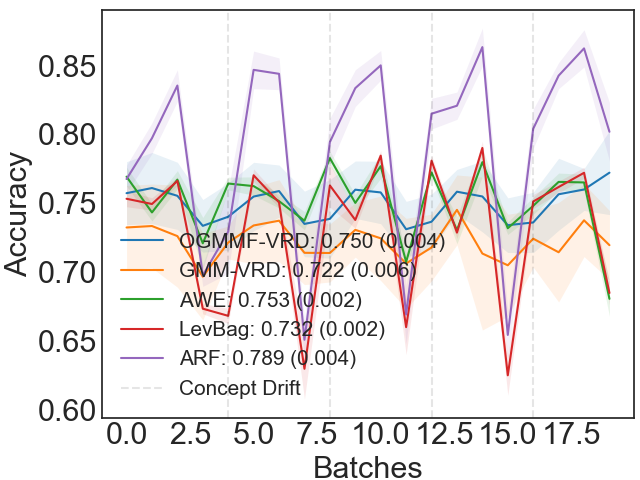}\vspace{-0.2cm}
        \caption{Sine 2 (Bests)}
        \label{fig:sine2 accuracy bests}
    \end{subfigure}
    ~ 
    \begin{subfigure}[h]{0.18\textwidth}
        \centering
        \includegraphics[height=1.05in]{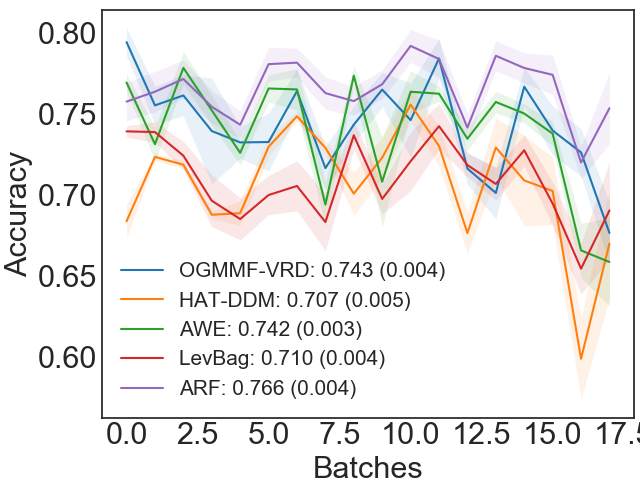}\vspace{-0.2cm}
        \caption{NOAA (Bests)}
        \label{fig:noaa accuracy bests}
    \end{subfigure}
    ~ 
    \begin{subfigure}[h]{0.18\textwidth}
        \centering
        \includegraphics[height=1.05in]{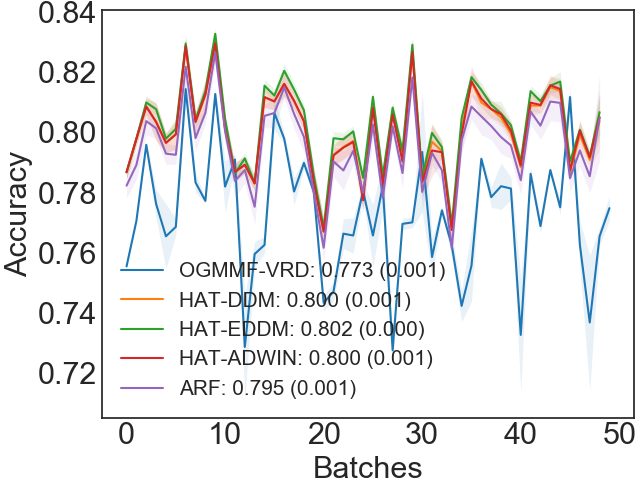}\vspace{-0.2cm}
        \caption{PAKDD (Bests)}
        \label{fig:pakdd accuracy bests}
    \end{subfigure}
    
    \caption{\textcolor{black}{Average accuracy over time for all methods on each dataset. The first row of figures (Figs. \ref{fig:virtual9 accuracy vr}, \ref{fig:circles accuracy vr}, \ref{fig:sine2 accuracy vr}, \ref{fig:noaa accuracy vr}, \ref{fig:pakdd accuracy vr}) shows only the methods that deal with virtual and real drifts (VR). The second row of figures (Figs. \ref{fig:virtual9 accuracy bests}, \ref{fig:circles accuracy bests}, \ref{fig:sine2 accuracy bests}, \ref{fig:noaa accuracy bests}, \ref{fig:pakdd accuracy bests}) shows only the five approaches with the best performances for each dataset (Bests).} The standard deviation is represented by shadow lines of the same color. Each point represents the accuracy for a batch observations, where 500 was used for synthetic datasets, and 1000 for real datasets.}
    \label{fig:accuracy over time}
\end{figure*}

\textcolor{black}{When comparing OGMMF-VRD to other approaches from the literature in terms of Accuracy (Fig. \ref{fig:friedman accuracy}), we observed that OGMMF-VRD was significantly better than the other methods, except for AWE and ARF, which are both ensemble-based methods. In terms of G-mean (Fig. \ref{fig:friedman gmean}), OGMMF-VRD was not statistically different only against AWE, ARF, and LevBag, which are also ensemble-based methods. With this, we observed that incremental learning methods combined with detectors performed poorly, which indicates that only incremental learning with a reset is not enough to tackle all types of drifts effectively.}
\textcolor{black}{In terms of execution time (Fig. \ref{fig:friedman time}), OGMMF-VRD was only statistically better than Dynse. This is because the proposed method performs many loops to (i) choose the best Gaussians to compose the GMM during training; and (ii) choosing the best GMMs to adapt when the drift happens. To understand these results better, we present in the second row of Fig. \ref{fig:accuracy over time}, the performance of the 5 best approaches for each dataset. The AOTs for other datasets are in the supplementary material.}

\textcolor{black}{The predictive performance of ensemble methods can be summarized in two main points based on the results obtained for the Sine 2 dataset (Fig. \ref{fig:sine2 accuracy bests}). The first point is that ensemble-based methods have a hard drop in their performance when drifts happen. This is because these methods were not proposed to understand the drift in order to apply a corresponding appropriate strategy, they just reset their knowledge. Although OGMMF-VRD also has a drop in its predictive performance, its recovery is much faster, which shows that the strategies for dealing with virtual and real drifts are efficient. The second point is related to the performance of the ensembles when there is no drift. We observed that the combination of several models allows better performance during periods of stability. In OGMMF-VRD, we use only a single classifier that ends up being inferior in terms of predictive performance compared to a combination of a group of models during such stable periods. Therefore, the combination of models can be an alternative in future work to further improve OGMMF-VRD's predictive performance.}

\subsection{Impact of Noise on Classifier Performance}
\label{subsec:noise analysis}

This experiment aims to determine how well the proposed approach answers RQ2 through its ability to deal with both real and virtual drifts while being robust to noise. To evaluate this, \textcolor{black}{we used all seven synthetic datasets discussed in Tbl. \ref{tbl:datasets} for the comparison with GMM-VRD, IGMM-CD, and Dynse, which are approaches that handle virtual and real drifts.} Thus we can see how the literature approaches discussed in Section \ref{subsec:comparison} behave in different types of drifts considering different noise levels as [5\%, 10\%, 15\%, 20\%]. The parameters used in the algorithms were the same as in the Tbl. \ref{tbl:parameters}. The noise represents the random exchange of X\% of the classes of the observations in the dataset. We include OGMMF-VRD with and without its noise filter mechanism  discussed in Section \ref{subsec:noise filter}, to understand how beneficial it is.

\begin{figure}[t]
    \centering
    \begin{subfigure}[h]{0.15\textwidth}
        \centering
        \includegraphics[height=1.in]{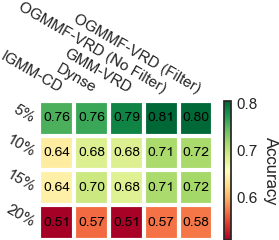}
        \caption{Virtual 9}
        \label{fig:virtual noise}
    \end{subfigure}
    ~ 
    \begin{subfigure}[h]{0.15\textwidth}
        \centering
        \includegraphics[height=1.0in]{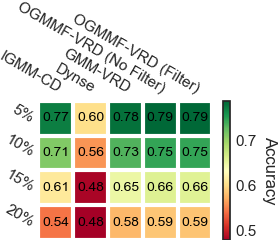}
        \caption{Sine 2}
        \label{fig:sine2 noise}
    \end{subfigure}
    ~ 
    \begin{subfigure}[h]{0.15\textwidth}
        \centering
        \includegraphics[height=1.0in]{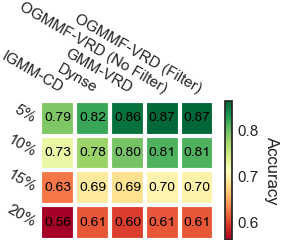}
        \caption{Circles}
        \label{fig:circles noise}
    \end{subfigure}
    
    \caption{\textcolor{black}{Average accuracy for only three of seven synthetic datasets with different noise levels. Cells closer to green/red represent better/worse results. These three datasets summarize the results obtained in all others. The other results are in the supplementary material.}} 
  \label{fig:noise results}
\end{figure}

\textcolor{black}{For reasons of space, in Fig. \ref{fig:noise results}, we present only three datasets (Virtual 9, Sine 2 and Circles) that summarize the results obtained for all the other seven. They have virtual drifts, real drifts, and both, respectively. The results for the other datasets are in the supplementary material.} We can see that, regardless of the approach, when the noise level is increased, the accuracy of all tends to decrease. This is due to the transformation of data into noise that does not represent any information to the classifiers.

To understand to what extent each approach is affected by noise, we perform the Friedman test \textcolor{black}{for all seven datasets considering all noise levels together in Fig. \ref{fig:friedman noise}. In the supplementary material, we performed different Friedman tests considering each noise level [5 \%, 10 \%, 15 \%, 20 \%] separately.} Based on the Nemenyi results in Fig. \ref{fig:friedman noise}, OGMMF-VRD with noise filter achieved the best results for average accuracy, while the IGMM-CD the worst. GMM-VRD performed poorly, showing that it is sensitive to noise, supporting our motivation for the noise filter. We found no evidence to reject the hypothesis that OGMMF-VRD with and without noise filter have the same overall accuracy \textcolor{black}{across datasets, even though the filter obtained a slightly higher ranking. This means that, for certain datasets, the noise filter may help, while for other datasets it may be detrimental. So, overall, across datasets, there was no significant difference between OGMMF-VRD with and without the noise filter.} 

\begin{figure}[t]
	\begin{center}
		\includegraphics[width=3.3in]{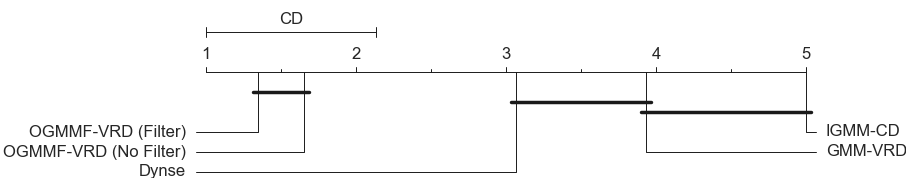}
	\end{center}
	\caption{\textcolor{black}{Friedman and Nemenyi tests for average accuracy for all seven synthetic datasets with all noise levels. Friedman's p-value was 1.04E-22} and its ranking is shown from left (best) to right (worst). Any pair of approaches whose distance between them is larger than CD is considered to be different.}
	\label{fig:friedman noise}
\end{figure}

\subsection{Impact of Proposed Approach's Mechanisms}
\label{subsec:mechanism analysis}

This experiment aims to complement the answer to RQ2 by checking whether it is really necessary to have the distinct virtual and real drift mechanisms. \textcolor{black}{It also checks how well the pool used by the proposed approach can harness past knowledge to accelerate adaptation to both virtual and real drifts, answering RQ3.} Only the synthetic datasets were used for this analysis, because they enable a better understanding of the behavior of the approaches in relation to each type of drift, unlike the real-world datasets. 

\begin{figure}[t]
    \centering
    \begin{subfigure}[h]{0.22\textwidth}
        \centering
        \includegraphics[height=1.23in]{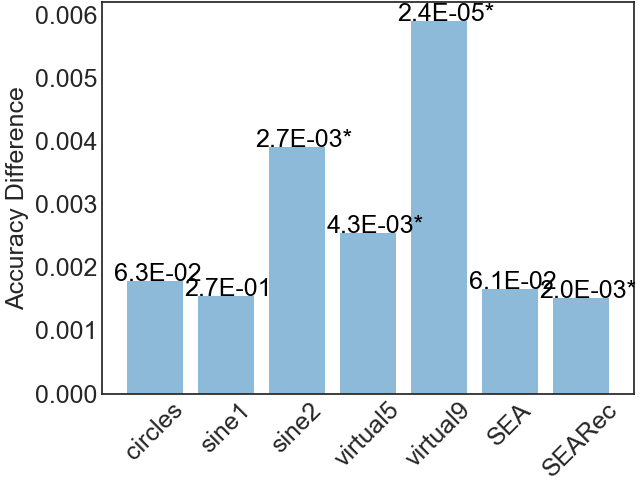}\vspace{-0.2cm}
        \caption{Bar: Virtual + Ns. Real}
        \label{fig:virtual part}
    \end{subfigure}
    ~ 
    \begin{subfigure}[h]{0.22\textwidth}
        \centering
        \includegraphics[height=1.19in]{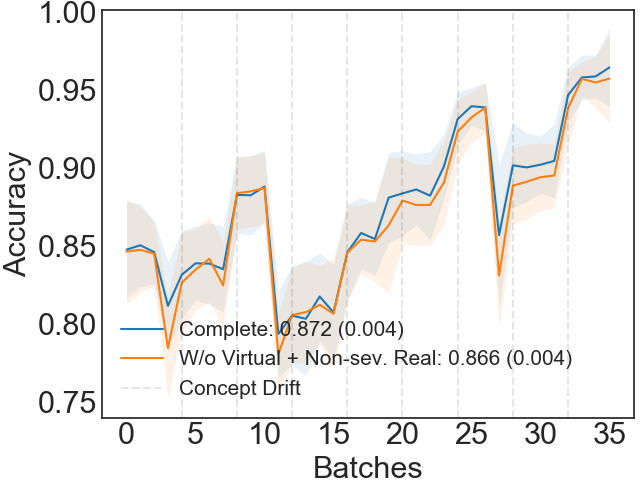}
        \caption{Bar: Virtual + Ns. Real}
        \label{fig:best virtual}
    \end{subfigure}
    ~ 
    \begin{subfigure}[h]{0.22\textwidth}
        \centering
        \includegraphics[height=1.23in]{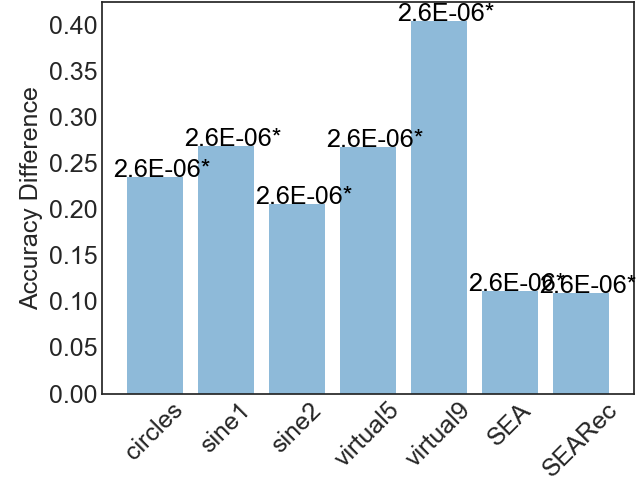}\vspace{-0.2cm}
        \caption{Bar: Severe Real}
        \label{fig:real part}
    \end{subfigure}
    ~ 
    \begin{subfigure}[h]{0.22\textwidth}
        \centering
        \includegraphics[height=1.19in]{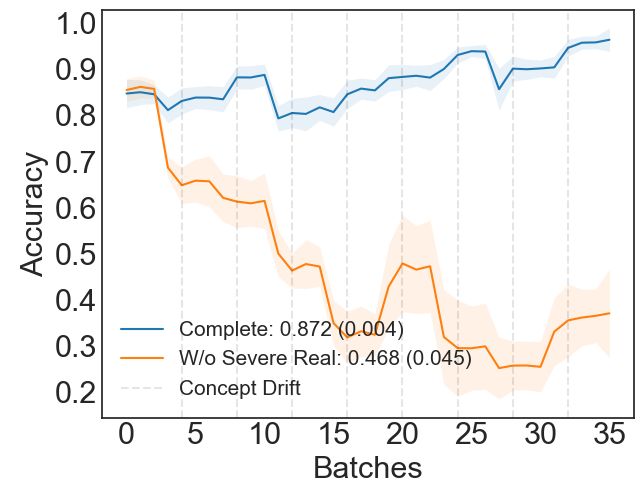}
        \caption{Best: Severe Real}
        \label{fig:best real}
    \end{subfigure}
    ~ 
    \begin{subfigure}[h]{0.22\textwidth}
        \centering
        \includegraphics[height=1.23in]{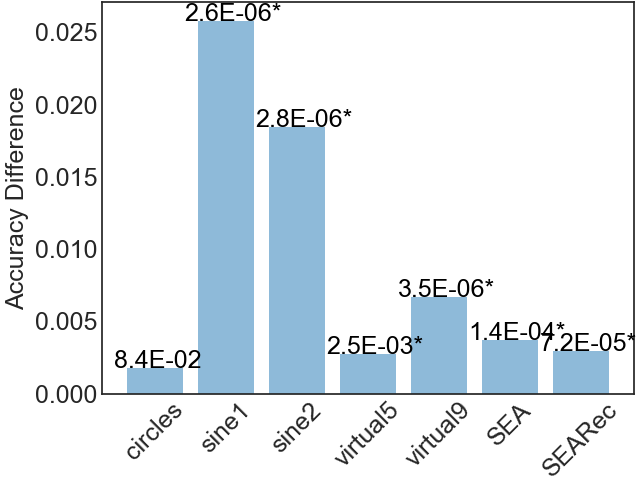}\vspace{-0.2cm}
        \caption{Bar: Pool}
        \label{fig:estimation part}
    \end{subfigure}
    ~ 
    \begin{subfigure}[h]{0.22\textwidth}
        \centering
        \includegraphics[height=1.19in]{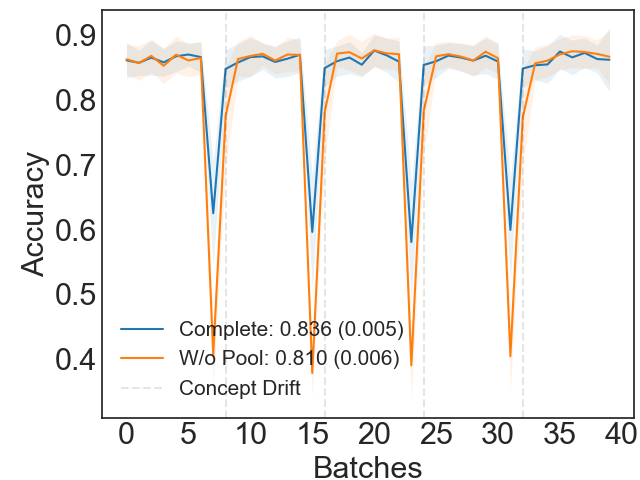}
        \caption{Best: Pool}
        \label{fig:best estimation}
    \end{subfigure}
    
    \caption{Accuracy improvements obtained by each of OGMM-VRD's mechanisms. Each bar represents the subtraction of the average of the complete system from the average of the system without a given mechanism (e.g. w/o non-severe drift adaptation). The number above each bar represents the p-value obtained by Wilcoxon test over the comparison pair to pair. P-values representing statistically significant difference at the level of $\alpha=0.05$ are marked by *. The AOT plots represent the dataset with the best improvement in each bar plot. The blue line represents the complete OGMMF-VRD system and the orange line represents the system without the respective mechanism.}
    
  \label{fig:mechanism wilcoxon}
\end{figure}

Fig. \ref{fig:virtual part}, and \ref{fig:real part}, \ref{fig:estimation part} present a bar chart of the improvement in accuracy obtained by the full OGMMF-VRD system w.r.t. the system without each given drift adaptation mechanism, \textcolor{black}{and above all bars we present the p-values of the corresponding Wilcoxon tests to support each pairwise comparison. For example, the average of the complete system minus the average of the system without the non-severe drift adaptation mechanism is in Fig. \ref{fig:virtual part}.} This experiment was made considering the parameters in Tbl. \ref{tbl:parameters} with $m=200$, because the more observations for training, the more the system as a whole is favored to obtain a robust model, thus allowing a better discussion of each mechanism's impact on the accuracy.



By analyzing the results for the virtual and non-severe real drift adaptation mechanism (Fig. \ref{fig:virtual part}), we observe that there was a statistically significant slight performance gain in 4 out of 7 datasets. Datasets with virtual drifts were the most favored. The SEAREc with the severity of 17.95-49\% also improved due to its gradual shift allowing the mechanism to track the drift before it becomes complete, despite this improvement being smaller than that of the datasets involving virtual drifts. Finally, Sine 2 also showed relatively large gains, indicating that this mechanism can also sometimes benefit the system in the presence of high severity real drifts. 
To visibly analyzing these gains we present in the Fig. \ref{fig:best virtual} the best gain obtained, which was for Virtual 9. In this dataset, changes occur in one class at a time, with observations appearing in another region of the space. This indicates that the proposed mechanism to create Gaussians can quickly prevent the system from losing performance. This complements the discussions of RQ2 in that the inclusion of a mechanism to deal with low severity drifts can help improving predictive performance in the presence of virtual drifts. 
  
By analyzing the results for the severe real drift adaptation mechanism (Fig. \ref{fig:real part}), we observe that this mechanism statistically significantly improved the performances in all cases, showing that this mechanism can be useful even for virtual drifts. This may be because CDTs can enable the system to quickly react to drifts as soon as they are detected. If the new concept is not too difficult to learn from scratch, enabling this quick reaction can potentially lead to faster adaptation than updating existing Gaussians depending on the adaptation parameters being used. This is unless the drifts have very low severity, in which case the mechanism for virtual and non-severe real drifts would likely still be the most helpful. 
In the best improvement that was for Virtual 9 (Fig. \ref{fig:best real}), we note that not using a CDT can be quite derogatory for model performance since the AOTs compared are very far apart. 

Finally, \textcolor{black}{we compared (i) GMM with EDDM without Pool against (ii) GMM with EDDM with Pool. Therefore, the analyzes concern how the use of the Pool improved the predictive performance of the system.} On the results for estimating GMMs from Pool in the presence of a concept drift (Fig. \ref{fig:estimation part}), we observed that 6 out of 7 cases significantly improved predictive performance. In SEARec, despite the gain is significant, the improvement was low due to the small pool size. The recurring drifts happen after 4 concepts, by which time the pool has been entirely renewed. The gains were mainly significant in datasets that have drifts that are both real, severe, abrupt and recurrent such as Sine 1 and Sine 2. Looking at the best case improvement, which was Sine 1 (Fig. \ref{fig:best estimation}), we see that in the presence of concept drifts, when estimating new models, system performance degrades less than when having to wait for a lot of data for re-initialize the classifier. These discussions support our RQ3, which states that harnessing the knowledge of past GMM's can accelerate the adaptation in both virtual and real drifts.  

\section{Conclusion}
\label{sec:conclusion}

This paper provides (i) a detailed understanding of the effects that virtual and real drifts have on classifiers’ suitability; (ii) the OGMMF-VRD approach, a system for dealing with virtual and real drift at the same time in classification data streams; and (iii) an unsupervised/supervised methodology with noise filter to train the GMM and achieve better robustness to noise.  

The computational experiments were performed to compare the OGMMF-VRD with existing approaches that claim to deal with both virtual and real drifts, \textcolor{black}{and other existing approaches that deal with concept drift in general}. The main results showed that the proposed approach achieved, in general, the best results in all metrics evaluated, \textcolor{black}{not being statistically different only to some ensemble-based methods.} We also showed that noise filter, pool, severe and non-severe adaptation mechanisms of the proposed OGMM-VRD performed well and usually lead to significant improvements for both virtual and real datasets. With these results, we answer the research questions of this work as follows:

\textbf{RQ1) What is the difference between the impact of real and virtual drifts on the suitability of classifiers' \textit{learned} decision boundaries and predictive performance over time?} (i) Due to the partial representation of the data, some types of classifiers learn incorrect decision boundaries while others based on GMM learn insufficient decision boundaries for the problem. For this reason, when a new observation from the trained class appears in non-trained regions, the classifiers make mistakes. (ii) Dealing with virtual drifts using the same strategy for real drifts wastes useful knowledge that could be used to expedite the classifier adaptation to the new concept especially in the case of GMM-based systems. (iii) In the experiments of mechanisms analysis, we saw that not incorporating a strategy to handle virtual drifts significantly drops the system's performance. (iv) Real drifts change the \textit{true} decision boundaries of the problem causing a significant drop in the classifier performance, but if the drift is non-severe, part of the knowledge can also be harnessed.

\textbf{RQ2) How to deal with both virtual and real drifts while achieving robustness to noise?} (i) Using a noise filter allowed us to avoid adapting to observations that would cause problems in our system. The experiments showed that this mechanism slightly improved the results compared to the literature approaches. (ii) Using pertinence threshold allowed us to create Gaussians quickly to properly deal with virtual drifts. (iii) Using \textit{on-line} learning has enabled us to maintain existing Gaussians by modeling them as new distributions arrived. (iv) Using a CDT to reset the system enabled us to swiftly trigger reaction to several drifts. (v) Experiments have shown that combinations of these techniques statistically improved system performance on both virtual and real drifts. 


\textbf{RQ3) How to best harness knowledge gained from similar concepts to accelerate adaptation to both virtual and real drifts?} (i) Saving older GMM's in a pool allowed the system to be able to choose the best classifier in the presence of similar concepts regardless the type of drift. (ii) The results showed that this mechanism led to statistically significant improvements on the system performance.



Although the results have shown the superior performance of  OGMMF-VRD, our proposals have some limitations, which should be addressed in future work: (i) it does not perform well on datasets with challenging class imbalanced distributions; (ii) it does not address the emergence of new classes; (iii) it discards useful knowledge when resetting the entire system; and (iv) it does not tackle verification latency. 

Future work includes a study of (i) how to adapt the mechanism of creating Gaussians over time to adapt the OGMMF-VRD to problems with class evolution; (ii) a study to do not learn from scratch the new concept, so we intend to monitor the error level of each Gaussian in the GMM in order to exclude only the degraded Gaussians; and (iii) to adjust the OGMMF-VRD to attendee real-world applications as software defect detection that has problems with class imbalance and the indefinite time for the arrival of the labels. 

\section*{Acknowledgment}

The authors would like to thank CNPq, FACEPE and CAPES (Grant No. 88887.588731/2020-00),  for their financial support. Leandro Minku was supported by EPSRC Grant No. EP/R006660/2.

\bibliographystyle{IEEEtran}
\bibliography{references}

\end{document}


\maketitle

\begin{table}[ht]
\caption{Code to generate the synthetic datasets with Tornado framework (Python) available on-line in https://github.com/alipsgh/tornado.}
\label{tbl:dataset generation}

\resizebox{\columnwidth}{!}{%
\begin{tabular}{|c|c|}
\hline
\textbf{Datasets} & \textbf{Code to Generate}                                                           \\ \hline
Circles           & CIRCLES(concept\_length=2000, noise\_rate=0.1)                                      \\ \hline
Sine1             & SINE1(concept\_length=2000, noise\_rate=0.1)                                        \\ \hline
Sine2             & SINE2(concept\_length=2000, noise\_rate=0.1)                                        \\ \hline
SEA               & SEA(concept\_length=2000, thresholds={[}1, 9, 2, 6{]}, noise\_rate=0.1)             \\ \hline
SEARec            & SEA(concept\_length=2000, thresholds={[}1, 9, 2, 6, 1, 9, 2, 6{]}, noise\_rate=0.1) \\ \hline
\end{tabular}
}
\end{table}

\begin{figure*}[ht]
    \centering
     \begin{subfigure}[h]{0.18\textwidth}
        \centering
        \includegraphics[height=1.05in]{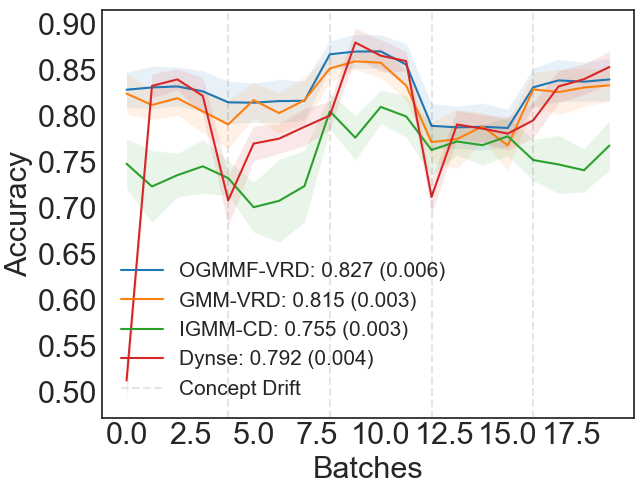}\vspace{-0.2cm}
        \caption{Virtual 5 drifts}
        \label{fig:virtual 5 changes accuracy}
    \end{subfigure}
    ~ 
    \begin{subfigure}[h]{0.18\textwidth}
        \centering
        \includegraphics[height=1.05in]{images/virtual9_vr_accuracy.png}\vspace{-0.2cm}
        \caption{Virtual 9 drifts}
        \label{fig:virtual 9 changes accuracy}
    \end{subfigure}
    ~
    \begin{subfigure}[h]{0.18\textwidth}
        \centering
        \includegraphics[height=1.05in]{images/circles_vr_accuracy.png}\vspace{-0.2cm}
        \caption{Circles}
        \label{fig:circles accuracy}
    \end{subfigure}
    ~ 
    \begin{subfigure}[h]{0.18\textwidth}
        \centering
        \includegraphics[height=1.05in]{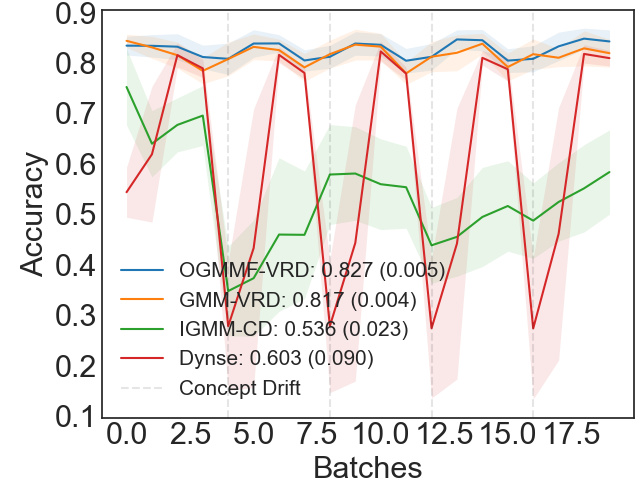}\vspace{-0.2cm}
        \caption{Sine1}
        \label{fig:sine1 accuracy}
    \end{subfigure}
    ~ 
    \begin{subfigure}[h]{0.18\textwidth}
        \centering
        \includegraphics[height=1.05in]{images/sine2_vr_accuracy.png}\vspace{-0.2cm}
        \caption{Sine2}
        \label{fig:sine2 accuracy}
    \end{subfigure}
    ~ 
    \begin{subfigure}[h]{0.18\textwidth}
        \centering
        \includegraphics[height=1.05in]{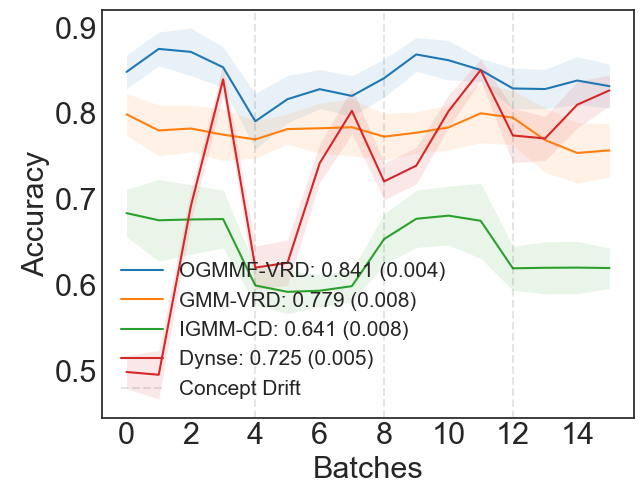}\vspace{-0.2cm}
        \caption{SEA}
        \label{fig:sea accuracy}
    \end{subfigure}
    ~ 
    \begin{subfigure}[h]{0.18\textwidth}
        \centering
        \includegraphics[height=1.05in]{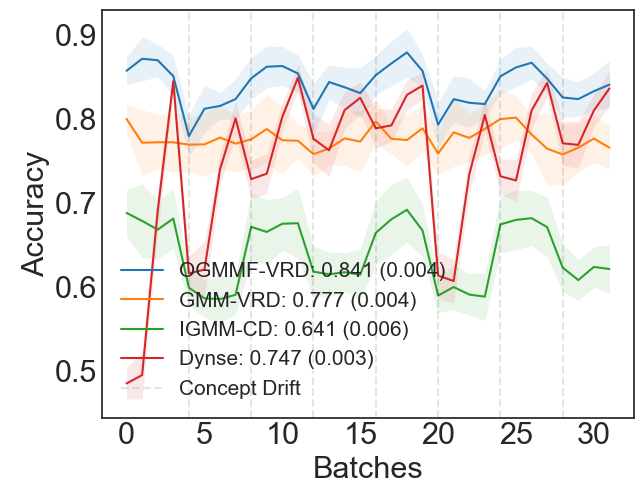}\vspace{-0.2cm}
        \caption{SEARec}
        \label{fig:searec accuracy}
    \end{subfigure}
    ~ 
    \begin{subfigure}[h]{0.18\textwidth}
        \centering
        \includegraphics[height=1.05in]{images/noaa_vr_accuracy.png}\vspace{-0.2cm}
        \caption{NOAA}
        \label{fig:noaa}
    \end{subfigure}
    ~ 
    \begin{subfigure}[h]{0.18\textwidth}
        \centering
        \includegraphics[height=1.05in]{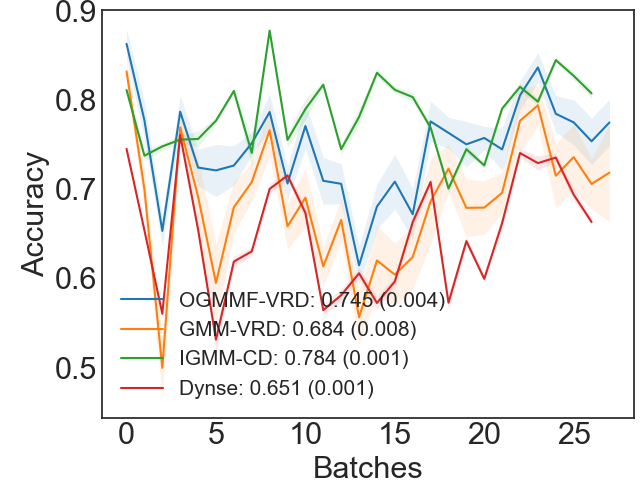}\vspace{-0.2cm}
        \caption{ELEC}
        \label{fig:elec}
    \end{subfigure}
    ~ 
    \begin{subfigure}[h]{0.18\textwidth}
        \centering
        \includegraphics[height=1.05in]{images/pakdd_vr_accuracy.png}\vspace{-0.2cm}
        \caption{PAKDD}
        \label{fig:pakdd}
    \end{subfigure}
    
    \caption{Mean of accuracy over time for methods that handle virtual and real drifts on each dataset. The standard deviation is represented by shadow lines of the same color. Each point represents the accuracy for a batch observations, where 500 was used for synthetic datasets, and 1000 for real datasets.}
    \label{fig:accuracy over time}
\end{figure*}

\begin{figure*}[ht]
    \centering
     \begin{subfigure}[h]{0.18\textwidth}
        \centering
        \includegraphics[height=1.05in]{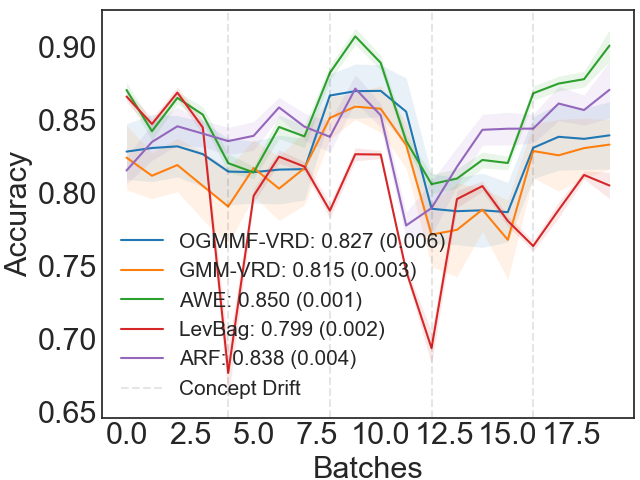}\vspace{-0.2cm}
        \caption{Virtual 5 drifts}
        \label{fig:virtual 5 changes accuracy}
    \end{subfigure}
    ~ 
    \begin{subfigure}[h]{0.18\textwidth}
        \centering
        \includegraphics[height=1.05in]{images/virtual9_best_accuracy.png}\vspace{-0.2cm}
        \caption{Virtual 9 drifts}
        \label{fig:virtual 9 changes accuracy}
    \end{subfigure}
    ~
    \begin{subfigure}[h]{0.18\textwidth}
        \centering
        \includegraphics[height=1.05in]{images/circles_best_accuracy.png}\vspace{-0.2cm}
        \caption{Circles}
        \label{fig:circles accuracy}
    \end{subfigure}
    ~ 
    \begin{subfigure}[h]{0.18\textwidth}
        \centering
        \includegraphics[height=1.05in]{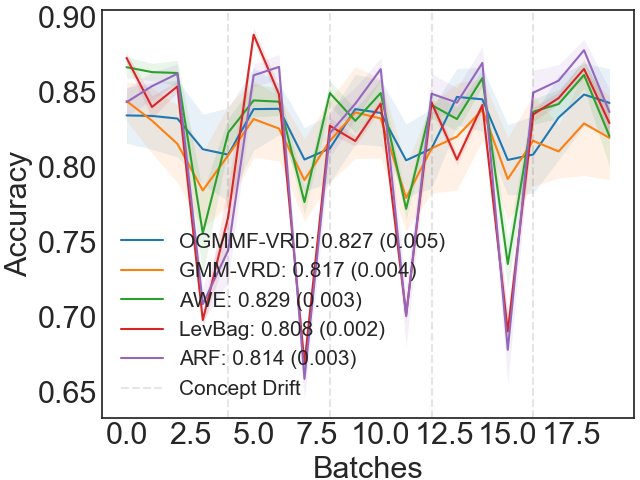}\vspace{-0.2cm}
        \caption{Sine1}
        \label{fig:sine1 accuracy}
    \end{subfigure}
    ~ 
    \begin{subfigure}[h]{0.18\textwidth}
        \centering
        \includegraphics[height=1.05in]{images/sine2_best_accuracy.png}\vspace{-0.2cm}
        \caption{Sine2}
        \label{fig:sine2 accuracy}
    \end{subfigure}
    ~ 
    \begin{subfigure}[h]{0.18\textwidth}
        \centering
        \includegraphics[height=1.05in]{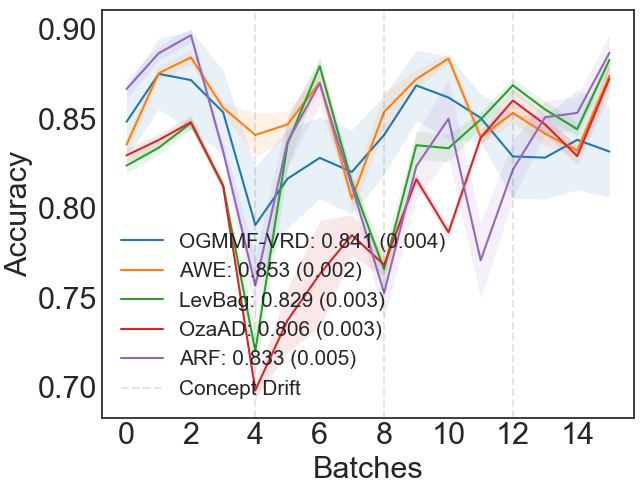}\vspace{-0.2cm}
        \caption{SEA}
        \label{fig:sea accuracy}
    \end{subfigure}
    ~ 
    \begin{subfigure}[h]{0.18\textwidth}
        \centering
        \includegraphics[height=1.05in]{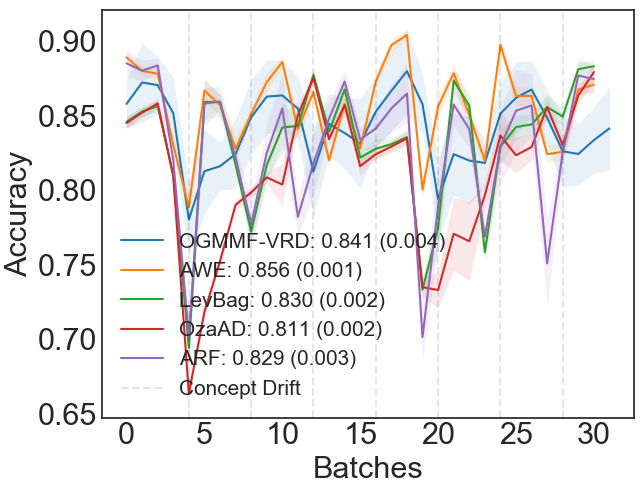}\vspace{-0.2cm}
        \caption{SEARec}
        \label{fig:searec accuracy}
    \end{subfigure}
    ~ 
    \begin{subfigure}[h]{0.18\textwidth}
        \centering
        \includegraphics[height=1.05in]{images/noaa_best_accuracy.png}\vspace{-0.2cm}
        \caption{NOAA}
        \label{fig:noaa}
    \end{subfigure}
    ~ 
    \begin{subfigure}[h]{0.18\textwidth}
        \centering
        \includegraphics[height=1.05in]{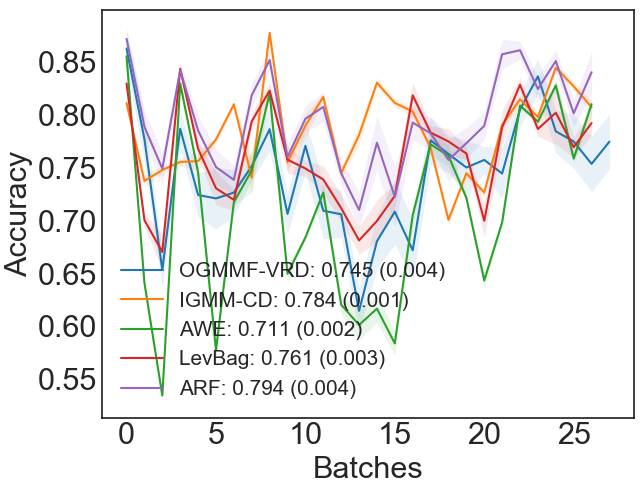}\vspace{-0.2cm}
        \caption{ELEC}
        \label{fig:elec}
    \end{subfigure}
    ~ 
    \begin{subfigure}[h]{0.18\textwidth}
        \centering
        \includegraphics[height=1.05in]{images/pakdd_best_accuracy.png}\vspace{-0.2cm}
        \caption{PAKDD}
        \label{fig:pakdd}
    \end{subfigure}
    
    \caption{Mean of accuracy over time for the five best methods in each dataset. The standard deviation is represented by shadow lines of the same color. Each point represents the accuracy for a batch observations, where 500 was used for synthetic datasets, and 1000 for real datasets.}
    \label{fig:accuracy over time}
\end{figure*}

\begin{table}[ht]
\caption{Accuracy for approaches compared for all experimented datasets. The results highlighted with bold represent the best accuracy in comparison with the other approaches.}
\label{tbl:accuracy}

\resizebox{\columnwidth}{!}{%
\begin{tabular}{|c|c|c|c|c|c|c|c|c|c|c|c|c|}
\hline
\textbf{Dataset} & \textbf{HAT-DDM} & \textbf{HAT-EDDM} & \textbf{HAT-ADWIN} & \textbf{AWE}            & \textbf{LevBag} & \textbf{OzaAD} & \textbf{OzaAS} & \textbf{ARF}            & \textbf{OGMMF-VRD} & \textbf{GMM-VRD} & \textbf{Dynse} & \textbf{IGMM-CD} \\ \hline
Circles          & 0,6189 (0,005)   & 0,4926 (0,003)    & 0,5592 (0,013)     & \textbf{0,8489 (0,001)} & 0,8122 (0,001)  & 0,7373 (0,008) & 0,7247 (0,001) & 0,815 (0,003)           & 0.813 (0.005)      & 0.787 (0.007)    & 0.741 (0.033)  & 0.609 (0.033)    \\ \hline
Sine1            & 0,5653 (0,025)   & 0,4975 (0,003)    & 0,5094 (0,005)     & \textbf{0,8286 (0,002)} & 0,8082 (0,001)  & 0,6584 (0,017) & 0,5658 (0,002) & 0,8137 (0,003)          & 0.827 (0.005)      & 0.817 (0.004)    & 0.603 (0.090)  & 0.536 (0.023)    \\ \hline
Sine2            & 0,7044 (0,003)   & 0,4978 (0,004)    & 0,6585 (0,013)     & 0,7532 (0,001)          & 0,7319 (0,002)  & 0,5893 (0,016) & 0,5525 (0,001) & \textbf{0,7886 (0,003)} & 0.750 (0.004)      & 0.722 (0.006)    & 0.584 (0.073)  & 0.538 (0.024)    \\ \hline
Virtual5         & 0,6941 (0,01)    & 0,3987 (0,002)    & 0,5127 (0,01)      & \textbf{0,8504 (0,001)} & 0,7987 (0,001)  & 0,7702 (0,005) & 0,7368 (0,002) & 0,838 (0,004)           & 0.827 (0.006)      & 0.815 (0.003)    & 0.792 (0.004)  & 0.755 (0.003)    \\ \hline
Virtual9         & 0,7082 (0,011)   & 0,4104 (0,002)    & 0,492 (0,014)      & \textbf{0,8697 (0,001)} & 0,8104 (0,004)  & 0,7658 (0,014) & 0,7267 (0,001) & 0,8451 (0,003)          & 0.858 (0.006)      & 0.845 (0.005)    & 0.805 (0.007)  & 0.804 (0.007)    \\ \hline
SEA              & 0,6975 (0,004)   & 0,5046 (0,003)    & 0,5233 (0,004)     & \textbf{0,8534 (0,001)} & 0,8295 (0,002)  & 0,8057 (0,003) & 0,8049 (0,001) & 0,8335 (0,004)          & 0.841 (0.004)      & 0.779 (0.008)    & 0.725 (0.004)  & 0.641 (0.008)    \\ \hline
SEARec           & 0,7146 (0,025)   & 0,5014 (0,002)    & 0,5279 (0,015)     & \textbf{0,8558 (0,001)} & 0,8304 (0,001)  & 0,8106 (0,002) & 0,8094 (0)     & 0,8291 (0,003)          & 0.841 (0.004)      & 0.777 (0.004)    & 0.747 (0.003)  & 0.641 (0.006)    \\ \hline
NOAA             & 0,7072 (0,004)   & 0,6862 (0,001)    & 0,6989 (0,007)     & 0,7416 (0,002)          & 0,7098 (0,003)  & 0,6634 (0,011) & 0,6869 (0,015) & \textbf{0,7656 (0,004)} & 0.743 (0.004)      & 0.708 (0.009)    & 0.462 (0.002)  & 0.420 (0.001)    \\ \hline
ELEC             & 0,6966 (0,013)   & 0,6752 (0,005)    & 0,6967 (0,006)     & 0,7112 (0,001)          & 0,7606 (0,003)  & 0,7059 (0,004) & 0,6785 (0,001) & \textbf{0,7935 (0,003)} & 0.745 (0.004)      & 0.685 (0.008)    & 0.651 (0.001)  & 0.783 (0.001)    \\ \hline
PAKDD            & 0,8002 (0,001)   & 0,8024 (0)        & 0,8001 (0,001)     & 0,6205 (0,041)          & 0,7109 (0,005)  & 0,3951 (0,026) & 0,5484 (0,037) & \textbf{0,7955 (0,001)} & 0.773 (0.001)      & 0.702 (0.016)    & 0.660 (0.002)  & 0.604 (0.007)    \\ \hline
\end{tabular}
}
\end{table}
\begin{table}[ht]
\caption{G-mean for approaches compared for all experimented datasets. The results highlighted with bold represent the best accuracy in comparison with the other approaches.}
\label{tbl:gmean}

\resizebox{\columnwidth}{!}{%
\begin{tabular}{|c|c|c|c|c|c|c|c|c|c|c|c|c|}
\hline
\textbf{Dataset} & \textbf{HAT-DDM} & \textbf{HAT-EDDM} & \textbf{HAT-ADWIN} & \textbf{AWE}           & \textbf{LevBag} & \textbf{OzaAD} & \textbf{OzaAS}         & \textbf{ARF}           & \textbf{OGMMF-VRD}     & \textbf{GMM-VRD} & \textbf{Dynse} & \textbf{IGMM-CD} \\ \hline
Circles          & 0.615 (0.008)    & 0.492 (0.004)     & 0.546 (0.019)      & 0.848 (0.002)          & 0.811 (0.002)   & 0.737 (0.009)  & 0.725 (0.002)          & 0.815 (0.004)          & 0.813 (0.005)          & 0.787 (0.007)    & 0.740 (0.033)  & 0.592 (0.041)    \\ \hline
Sine1            & 0.562 (0.032)    & 0.496 (0.004)     & 0.498 (0.012)      & \textbf{0.829 (0.003)} & 0.808 (0.002)   & 0.658 (0.029)  & 0.561 (0.006)          & 0.814 (0.003)          & 0.827 (0.005)          & 0.817 (0.004)    & 0.603 (0.090)  & 0.521 (0.041)    \\ \hline
Sine2            & 0.704 (0.004)    & 0.497 (0.004)     & 0.657 (0.017)      & 0.753 (0.002)          & 0.732 (0.002)   & 0.588 (0.020)  & 0.548 (0.003)          & \textbf{0.789 (0.004)} & 0.749 (0.004)          & 0.722 (0.006)    & 0.584 (0.073)  & 0.530 (0.028)    \\ \hline
Virtual5         & 0.691 (0.013)    & 0.370 (0.005)     & 0.483 (0.017)      & \textbf{0.844 (0.002)} & 0.788 (0.002)   & 0.766 (0.006)  & 0.733 (0.002)          & 0.836 (0.004)          & 0.823 (0.006)          & 0.809 (0.004)    & 0.785 (0.005)  & 0.746 (0.004)    \\ \hline
Virtual9         & 0.707 (0.012)    & 0.404 (0.003)     & 0.474 (0.024)      & \textbf{0.866 (0.001)} & 0.807 (0.006)   & 0.761 (0.018)  & 0.724 (0.001)          & 0.843 (0.004)          & 0.855 (0.006)          & 0.842 (0.006)    & 0.802 (0.007)  & 0.799 (0.010)    \\ \hline
SEA              & 0.669 (0.009)    & 0.496 (0.006)     & 0.302 (0.090)      & \textbf{0.853 (0.002)} & 0.828 (0.003)   & 0.805 (0.003)  & 0.805 (0.001)          & 0.833 (0.005)          & 0.841 (0.004)          & 0.779 (0.008)    & 0.725 (0.005)  & 0.633 (0.008)    \\ \hline
SEARec           & 0.712 (0.027)    & 0.501 (0.003)     & 0.489 (0.044)      & \textbf{0.856 (0.001)} & 0.830 (0.002)   & 0.811 (0.002)  & 0.809 (0.001)          & 0.829 (0.003)          & 0.841 (0.004)          & 0.777 (0.004)    & 0.747 (0.003)  & 0.632 (0.006)    \\ \hline
NOAA             & 0.457 (0.053)    & 0.121 (0.028)     & 0.367 (0.070)      & 0.683 (0.006)          & 0.708 (0.003)   & 0.680 (0.005)  & 0.662 (0.020)          & 0.654 (0.009)          & \textbf{0.697 (0.005)} & 0.679 (0.015)    & 0.589 (0.003)  & 0.596 (0.005)    \\ \hline
ELEC             & 0.644 (0.021)    & 0.608 (0.010)     & 0.642 (0.012)      & 0.686 (0.002)          & 0.740 (0.004)   & 0.674 (0.009)  & 0.569 (0.004)          & \textbf{0.777 (0.004)} & 0.729 (0.005)          & 0.670 (0.009)    & 0.618 (0.002)  & 0.785 (0.001)    \\ \hline
PAKDD            & 0.070 (0.014)    & 0.014 (0.006)     & 0.071 (0.016)      & 0.551 (0.030)          & 0.449 (0.014)   & 0.482 (0.028)  & \textbf{0.557 (0.030)} & 0.189 (0.008)          & 0.240 (0.003)          & 0.419 (0.013)    & 0.518 (0.001)  & 0.475 (0.001)    \\ \hline
\end{tabular}
}
\end{table}
\begin{table}[ht]
\caption{Time for approaches compared for all experimented datasets. The results highlighted with bold represent the best accuracy in comparison with the other approaches.}
\label{tbl:time}

\resizebox{\columnwidth}{!}{%
\begin{tabular}{|c|c|c|c|c|c|c|c|c|c|c|c|c|}
\hline
\textbf{Datasets} & \textbf{HAT-DDM} & \textbf{HAT-EDDM} & \textbf{HAT-ADWIN}    & \textbf{AWE} & \textbf{LevBag} & \textbf{OzaAD} & \textbf{OzaAS}       & \textbf{ARF} & \textbf{OGMMF-VRD} & \textbf{GMM-VRD} & \textbf{Dynse} & \textbf{IGMM-CD} \\ \hline
Circles           & 16,8 (0,09)      & 17,1 (0,07)       & 16,7 (0,1)            & 18,9 (0,07)  & 20,6 (0,06)     & 18,5(0,07)     & \textbf{16,1 (0,08)} & 27,6 (0,2)   & 140.9 (44.6)       & 55.6 (14.6)      & 733.5 (56.8)   & 221.2 (90.9)     \\ \hline
Sine1             & 25,5 (0,1)       & 25,4 (0,1)        & 24,8 (0,15)           & 27,9 (0,1)   & 29,9 (0,16)     & 27,2 (0,1)     & \textbf{24,3 (0,1)}  & 36,5 (0,2)   & 205.6 (68.6)       & 78.4 (24.6)      & 1078.7 (33.7)  & 249.3 (86.2)     \\ \hline
Sine2             & 25,5 (0,09)      & 25,4 (0,08)       & 25,4 (0,12)           & 27,9 (0,1)   & 29,8 (0,1)      & 27,3 (0,1)     & \textbf{24,3 (0,09)} & 36,6 (0,1)   & 237.4 (85.2)       & 84.5 (25.5)      & 1048.6 (38.2)  & 258.4 (87.1)     \\ \hline
Virtual5          & 25,1 (0,2)       & 24,5 (0,09)       & 23,9 (0,09)           & 27,8 (0,07)  & 29,1 (0,09)     & 26,6 (0,09)    & \textbf{23,3 (0,1)}  & 38,6 (0,5)   & 346.6 (121.2)      & 102.4 (25.7)     & 881.2 (63.5)   & 648.3 (193.1)    \\ \hline
Virtual9          & 20,8 (0,1)       & 20,3 (0,1)        & 19,7 (0,13)           & 23,2 (0,09)  & 24,4 (0,1)      & 22,1 (0,1)     & \textbf{19,2 (0,06)} & 32,6 (0,6)   & 219.9 (62.4)       & 75.2 (20.5)      & 839.6 (61.1)   & 499.4 (162.6)    \\ \hline
SEA               & 17,3 (0,1)       & 17,1 (0,06)       & 16,5 (0,07)           & 19,8 (0,05)  & 21,2 (0,06)     & 18,9 (0,06)    & \textbf{16,3 (0,1)}  & 28,5 (0,3)   & 149.5 (61.4)       & 60.7 (19.8)      & 589.5 (50.1)   & 233.6 (88.3)     \\ \hline
SEARec            & 62,5 (0,5)       & 60,9 (0,3)        & 60,1 (0,4)            & 66,9 (0,2)   & 69,5 (0,2)      & 64,9 (0,2)     & \textbf{59,6 (0,06)} & 86,3 (1,3)   & 385.4 (117.3)      & 166.6 (55.3)     & 1099.7 (81.7)  & 549.7 (183.9)    \\ \hline
NOAA              & 76,8 (1,2)       & 76,4 (1,2)        & 75,8 (1,3)            & 88,5 (1,3)   & 90,4 (1,5)      & 81,9 (1,2)     & \textbf{74,6 (1,1)}  & 120,5 (3,7)  & 987.3 (354.9)      & 367.1 (119.7)    & 2495.7 (125.5) & 1099.1 (367.8)   \\ \hline
ELEC              & 165,6 (0,9)      & 165,4 (0,8)       & \textbf{164,7 (0,8)}  & 178,5 (0,8)  & 183,4 (0,9)     & 174,6 (0,6)    & 165,1 (0,6)          & 212,1 (2,5)  & 1122.2 (69.1)      & 528.6 (67.9)     & 3383.5 (43.6)  & 1202.8 (215.6)   \\ \hline
PAKDD             & 548,6 (16,9)     & 533,8 (12,1)      & \textbf{549,7 (14,9)} & 667,5 (16,8) & 652,3 (16,8)    & 596,1 (15,9)   & 559,6 (16,1)         & 749,2 (35,2) & 3997.9 (1476.1)    & 2498.1 (2244.3)  & 7285.5 (61.1)  & 2858.6 (1085.1)  \\ \hline
\end{tabular}
}
\end{table}

\begin{table}[ht]
\caption{Grid search for radius parameter of the OGMMF-VRD. The results highlighted with bold represent the best accuracy. According to the p-value 4.21E-01 obtained by the Friedman test, the H0 was not rejected, which means that this parameter did not have significant impact on the models' accuracy.}
\label{tbl:cfc}

\resizebox{\columnwidth}{!}{%
\begin{tabular}{|c|c|c|c|c|}
\hline
\textbf{Datasets} & \textbf{radius=10}         & \textbf{radius=15}         & \textbf{radius=20}         & \textbf{radius=25} \\ \hline
Circles           & 0,8143 (0,003)          & \textbf{0,8145 (0,003)} & 0,813 (0,004)           & 0,8136 (0,003)  \\ \hline
Sine1             & 0,8225 (0,003)          & 0,823 (0,004)           & \textbf{0,8239 (0,003)} & 0,8224 (0,003)  \\ \hline
Sine2             & 0,7475 (0,003)          & \textbf{0,7482 (0,004)} & 0,7475 (0,003)          & 0,7481 (0,002)  \\ \hline
Virtual5          & \textbf{0,8299 (0,003)} & 0,8295 (0,003)          & 0,8298 (0,003)          & 0,8292 (0,003)  \\ \hline
Virtual9          & 0,8555 (0,003)          & 0,8543 (0,003)          & \textbf{0,8569 (0,004)} & 0,8558 (0,002)  \\ \hline
SEA               & 0,8415 (0,003)          & 0,841 (0,005)           & \textbf{0,8436 (0,004)} & 0,8417 (0,003)  \\ \hline
\end{tabular}
}
\end{table}
\begin{table}[ht]
\caption{Grid search for EDDM drift level threshold used in OGMMF-VRD. The results highlighted with bold represent the best accuracy. According to the p-value 8.29E-01 obtained by the Friedman test, the H0 was not rejected, which means that this parameter did not have significant impact on the models' accuracy.}
\label{tbl:drift level}

\resizebox{\columnwidth}{!}{%
\begin{tabular}{|c|c|c|c|c|}
\hline
\textbf{Datasets} & \textbf{c=1}            & \textbf{c=1.5}          & \textbf{c=2}            & \textbf{c=2.5}          \\ \hline
Circles           & \textbf{0.8138 (0.003)} & 0.8133 (0.004)          & 0.813 (0.004)           & 0.8129 (0.004)          \\ \hline
Sine1             & 0.8221 (0.003)          & 0.8219 (0.002)          & \textbf{0.8227 (0.003)} & 0.8212 (0.003)          \\ \hline
Sine2             & 0.7478 (0.003)          & 0.7467 (0.003)          & \textbf{0.7488 (0.003)} & 0.7478 (0.003)          \\ \hline
Virtual5          & \textbf{0.8293 (0.004)} & 0.8287 (0.003)          & 0.8288 (0.003)          & 0.8295 (0.003)          \\ \hline
Virtual9          & 0.8563 (0.004)          & \textbf{0.8573 (0.003)} & 0.8564 (0.003)          & 0.8564 (0.004)          \\ \hline
SEA               & 0.8416 (0.005)          & \textbf{0.8416 (0.004)} & 0.8408 (0.004)          & \textbf{0.8416 (0.004)} \\ \hline
\end{tabular}
}
\end{table}

\begin{figure*}[ht]
    \begin{subfigure}[h]{1.1\textwidth}
        \centering
        \includegraphics[height=1.35in]{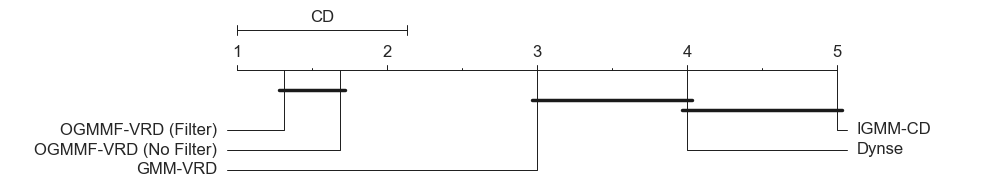}
        \caption{5\%}
        \label{fig:noisy 5}
    \end{subfigure}
    ~
    \begin{subfigure}[h]{1.1\textwidth}
        \centering
        \includegraphics[height=1.35in]{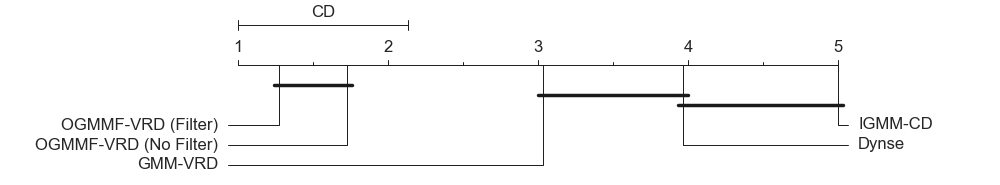}
        \caption{10\%}
        \label{fig:noisy 10}
    \end{subfigure}
    ~
    \begin{subfigure}[h]{1.1\textwidth}
        \centering
        \includegraphics[height=1.35in]{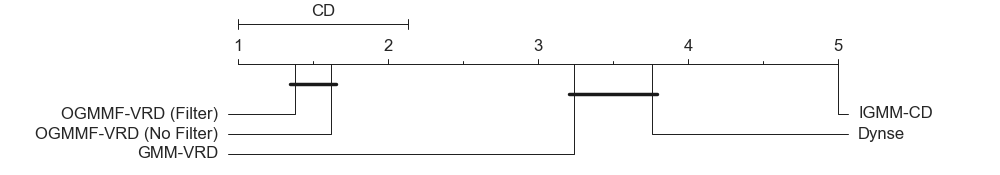}
        \caption{15\%}
        \label{fig:noisy 15}
    \end{subfigure}
    ~
    \begin{subfigure}[h]{1.1\textwidth}
        \centering
        \includegraphics[height=1.35in]{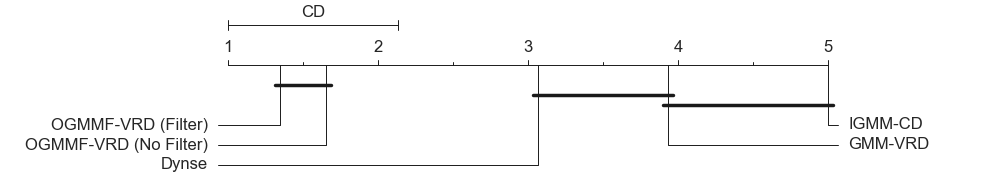}
        \caption{20\%}
        \label{fig:noisy 20}
    \end{subfigure}
   
    \caption{Friedman and Nemenyi tests for average accuracy for synthetic datasets with different noise levels. Friedman's p-value were p=4.38E-23, p=5.44E-23, p=4.50E-22, p=1.04E-22 and its ranking is shown from left (best) to right (worst). Any pair of approaches whose distance between them is larger than CD is considered to be significantly different.}
    \label{fig:noisy tests}
\end{figure*}

\begin{table}[ht]
\caption{Average accuracy for synthetic datasets with different noise levels. The results highlighted with bold represent the best accuracy in comparison with the other approaches.}
\label{tbl:noises}

\resizebox{\columnwidth}{!}{%
\begin{tabular}{|c|c|c|c|c|c|}
\hline
\textbf{Datasets} & \textbf{OGMMF-VRD (Filter)} & \textbf{OGMMF-VRD (No Filter)} & \textbf{GMM-VRD} & \textbf{Dynse} & \textbf{IGMM-CD} \\ \hline
Circles 5\%       & 0,8652 (0,003)              & \textbf{0,8657 (0,004)}        & 0,8583 (0,004)   & 0,8227 (0,013) & 0,7923 (0,003)   \\ \hline
Circles 10\%      & \textbf{0,8143 (0,003)}     & 0,814 (0,004)                  & 0,798 (0,004)    & 0,78 (0,011)   & 0,7291 (0,002)   \\ \hline
Circles 15\%      & 0,703 (0,004)               & \textbf{0,7043 (0,004)}        & 0,6854 (0,004)   & 0,6944 (0,014) & 0,6276 (0,003)   \\ \hline
Circles 20\%      & \textbf{0,6118 (0,003)}     & 0,611 (0,004)                  & 0,5991 (0,006)   & 0,6124 (0,009) & 0,5619 (0,003)   \\ \hline
Sine1 5\%         & 0,889 (0,005)               & \textbf{0,8902 (0,003)}        & 0,8519 (0,002)   & 0,7791 (0,001) & 0,7628 (0,003)   \\ \hline
Sine1 10\%        & \textbf{0,8313 (0,004)}     & 0,8295 (0,004)                 & 0,8121 (0,003)   & 0,7445 (0,001) & 0,7042 (0,001)   \\ \hline
Sine1 15\%        & \textbf{0,8313 (0,004)}     & 0,8295 (0,004)                 & 0,8121 (0,003)   & 0,7445 (0,001) & 0,7042 (0,001)   \\ \hline
Sine1 20\%        & \textbf{0,7157 (0,003)}     & 0,715 (0,004)                  & 0,7207 (0,003)   & 0,6738 (0,001) & 0,6156 (0,002)   \\ \hline
Sine2 5\%         & \textbf{0,7888 (0,003)}     & 0,7888 (0,003)                 & 0,7821 (0,003)   & 0,4781 (0,026) & 0,7741 (0,002)   \\ \hline
Sine2 10\%        & 0,7475 (0,003)              & \textbf{0,7476 (0,003)}        & 0,7349 (0,003)   & 0,4819 (0,024) & 0,7133 (0,002)   \\ \hline
Sine2 15\%        & 0,658 (0,004)               & \textbf{0,6601 (0,004)}        & 0,6482 (0,003)   & 0,5602 (0,093) & 0,612 (0,002)    \\ \hline
Sine2 20\%        & \textbf{0,5856 (0,003)}     & \textbf{0,587 (0,002)}         & 0,5788 (0,004)   & 0,6024 (0,002) & 0,5449 (0,003)   \\ \hline
Virtual5 5\%      & \textbf{0,8018 (0,004)}     & 0,8014 (0,003)                 & 0,8076 (0,002)   & 0,7575 (0,001) & 0,7176 (0,001)   \\ \hline
Virtual5 10\%     & \textbf{0,7243 (0,003)}     & 0,724 (0,004)                  & 0,7309 (0,006)   & 0,6894 (0,001) & 0,6236 (0,001)   \\ \hline
Virtual5 15\%     & 0,6226 (0,003)              & \textbf{0,6257 (0,003)}        & 0,6349 (0,006)   & 0,5992 (0,001) & 0,5228 (0,002)   \\ \hline
Virtual5 20\%     & \textbf{0,5189 (0,004)}     & 0,5167 (0,004)                 & 0,5196 (0,005)   & 0,4904 (0,001) & 0,4357 (0,001)   \\ \hline
Virtual9 5\%      & \textbf{0,8042 (0,003)}     & 0,806 (0,004)                  & 0,7879 (0,006)   & 0,7641 (0,026) & 0,7568 (0,002)   \\ \hline
Virtual9 10\%     & \textbf{0,7153 (0,004)}     & 0,7143 (0,003)                 & 0,6843 (0,007)   & 0,6849 (0,022) & 0,6413 (0,002)   \\ \hline
Virtual9 15\%     & \textbf{0,7177 (0,004)}     & 0,7145 (0,004)                 & 0,6836 (0,008)   & 0,6958 (0,004) & 0,6449 (0,002)   \\ \hline
Virtual9 20\%     & \textbf{0,5752 (0,003)}     & 0,572 (0,004)                  & 0,5141 (0,012)   & 0,5689 (0,002) & 0,5108 (0,001)   \\ \hline
SEA 5\%           & \textbf{0,9133 (0,002)}     & 0,9118 (0,002)                 & 0,9052 (0,002)   & 0,8718 (0,001) & 0,6476 (0,001)   \\ \hline
SEA 10\%          & 0,8555 (0,002)              & \textbf{0,8576 (0,002)}        & 0,8487 (0,005)   & 0,8221 (0,001) & 0,6103 (0,001)   \\ \hline
SEA 15\%          & \textbf{0,798 (0,003)}      & 0,7992 (0,004)                 & 0,7911 (0,003)   & 0,7717 (0,001) & 0,5885 (0,001)   \\ \hline
SEA 20\%          & 0,7458 (0,004)              & \textbf{0,7474 (0,002)}        & 0,7475 (0,007)   & 0,7275 (0,001) & 0,5687 (0,001)   \\ \hline
SEARec 5\%        & 0,9115 (0,002)              & \textbf{0,9128 (0,001)}        & 0,9022 (0,004)   & 0,8671 (0)     & 0,635 (0,001)    \\ \hline
SEARec 10\%       & 0,8519 (0,002)              & \textbf{0,8541 (0,001)}        & 0,8307 (0,005)   & 0,8197 (0,001) & 0,6066 (0,002)   \\ \hline
SEARec 15\%       & \textbf{0,7924 (0,003)}     & 0,7943 (0,002)                 & 0,7766 (0,004)   & 0,7733 (0,001) & 0,5859 (0,001)   \\ \hline
SEARec 20\%       & \textbf{0,7405 (0,002)}     & 0,7399 (0,002)                 & 0,7224 (0,003)   & 0,7289 (0,001) & 0,5611 (0,001)   \\ \hline
\end{tabular}
}
\end{table}
\begin{table}[ht]
\caption{Accuracy improvements obtained by each of OGMMF-VRD’s mechanisms. Each column represents OGMMF-VRD without a given mechanism (e.g. w/o non-severe drift adaptation). P-values representing statistically significant difference at the level of $\alpha$ = 0:05 are marked by *. The results highlighted with bold represent the best accuracy.}
\label{tbl:cfc}

\resizebox{\columnwidth}{!}{%
\begin{tabular}{|c|c|c|c|c|}
\hline
\textbf{Datasets} & \textbf{W/0 Virtual + Ns Real} & \textbf{W/0 Severe Real} & \textbf{W/0 Pool} & \textbf{OGMMF-VRD}      \\ \hline
Circles           & 0,8313 (0,005)                 & 0,5983 (0,012)*          & 0,8314 (0,003)    & \textbf{0,8332 (0,004)} \\ \hline
Sine1             & 0,8341 (0,004)                 & 0,567 (0,004)*           & 0,8098 (0,005)*   & \textbf{0,8356 (0,005)} \\ \hline
Sine2             & 0,7533 (0,003)*                & 0,5519 (0,004)*          & 0,739 (0,004)*    & \textbf{0,757 (0,004)}  \\ \hline
Virtual5          & 0,841 (0,003)*                 & 0,5762 (0,014)*          & 0,8408 (0,003)*   & \textbf{0,8436 (0,003)} \\ \hline
Virtual9          & 0,8662 (0,004)*                & 0,4684 (0,022)*          & 0,8656 (0,004)*   & \textbf{0,8721 (0,004)} \\ \hline
SEA               & 0,8606 (0,003)                 & 0,7501 (0,025)*          & 0,8588 (0,003)*   & \textbf{0,8625 (0,003)} \\ \hline
SEARec            & 0,86 (0,002)*                  & 0,7499 (0,014)*          & 0,858 (0,002)*    & \textbf{0,8611 (0,002)} \\ \hline
\end{tabular}
}
\end{table}

\begin{longtable}{|c|c|c|c|c|}
\caption{Recall for approaches compared for all experimented datasets. The results highlighted with bold represent the best accuracy in comparison with the other approaches.}
\label{tbl:recall}\\

\hline
\textbf{Dataset}           & \textbf{Approaches} & \textbf{Class 0}       & \textbf{Class 1}       & \textbf{Class 2}       \\ \hline
\multirow{12}{*}{Circles}  & HAT-DDM             & 0.580 (0.054)          & 0.657 (0.050)          & -                      \\ \cline{2-5} 
                           & HAT-EDDM            & 0.482 (0.030)          & 0.503 (0.028)          & -                      \\ \cline{2-5} 
                           & HAT-ADWIN           & 0.528 (0.104)          & 0.590 (0.128)          & -                      \\ \cline{2-5} 
                           & AWE                 & \textbf{0.810 (0.003)} & \textbf{0.888 (0.002)} & -                      \\ \cline{2-5} 
                           & LevBag              & 0.771 (0.003)          & 0.853 (0.003)          & -                      \\ \cline{2-5} 
                           & OzaAD               & 0.742 (0.008)          & 0.733 (0.012)          & -                      \\ \cline{2-5} 
                           & OzaAS               & 0.731 (0.005)          & 0.718 (0.003)          & -                      \\ \cline{2-5} 
                           & ARF                 & 0.817 (0.004)          & 0.813 (0.005)          & -                      \\ \cline{2-5} 
                           & OGMMF-VRD           & 0.806 (0.010)          & 0.820 (0.009)          & -                      \\ \cline{2-5} 
                           & GMM-VRD             & 0.754 (0.021)          & 0.821 (0.012)          & -                      \\ \cline{2-5} 
                           & Dynse               & 0.752 (0.015)          & 0.729 (0.055)          & -                      \\ \cline{2-5} 
                           & IGMM-CD             & 0.659 (0.115)          & 0.558 (0.146)          & -                      \\ \hline
\multirow{12}{*}{Sine1}    & HAT-DDM             & 0.600 (0.067)          & 0.531 (0.063)          & -                      \\ \cline{2-5} 
                           & HAT-EDDM            & 0.522 (0.023)          & 0.473 (0.023)          & -                      \\ \cline{2-5} 
                           & HAT-ADWIN           & 0.597 (0.067)          & 0.422 (0.065)          & -                      \\ \cline{2-5} 
                           & AWE                 & 0.835 (0.005)          & 0.822 (0.004)          & -                      \\ \cline{2-5} 
                           & LevBag              & 0.809 (0.002)          & 0.808 (0.002)          & -                      \\ \cline{2-5} 
                           & OzaAD               & 0.666 (0.031)          & 0.651 (0.032)          & -                      \\ \cline{2-5} 
                           & OzaAS               & 0.638 (0.031)          & 0.494 (0.033)          & -                      \\ \cline{2-5} 
                           & ARF                 & 0.809 (0.004)          & 0.818 (0.004)          & -                      \\ \cline{2-5} 
                           & OGMMF-VRD           & 0.826 (0.009)          & \textbf{0.827 (0.007)} & -                      \\ \cline{2-5} 
                           & GMM-VRD             & \textbf{0.815 (0.009)} & 0.819 (0.003)          & -                      \\ \cline{2-5} 
                           & Dynse               & 0.604 (0.093)          & 0.602 (0.088)          & -                      \\ \cline{2-5} 
                           & IGMM-CD             & 0.534 (0.121)          & 0.538 (0.125)          & -                      \\ \hline
\multirow{12}{*}{Sine2}    & HAT-DDM             & 0.699 (0.015)          & 0.709 (0.016)          & -                      \\ \cline{2-5} 
                           & HAT-EDDM            & 0.483 (0.016)          & 0.513 (0.016)          & -                      \\ \cline{2-5} 
                           & HAT-ADWIN           & 0.631 (0.040)          & 0.686 (0.029)          & -                      \\ \cline{2-5} 
                           & AWE                 & 0.750 (0.002)          & 0.756 (0.003)          & -                      \\ \cline{2-5} 
                           & LevBag              & 0.726 (0.003)          & 0.737 (0.002)          & -                      \\ \cline{2-5} 
                           & OzaAD               & 0.554 (0.021)          & 0.624 (0.022)          & -                      \\ \cline{2-5} 
                           & OzaAS               & 0.491 (0.015)          & 0.613 (0.013)          & -                      \\ \cline{2-5} 
                           & ARF                 & \textbf{0.785 (0.005)} & \textbf{0.792 (0.005)} & -                      \\ \cline{2-5} 
                           & OGMMF-VRD           & 0.752 (0.008)          & 0.748 (0.008)          & -                      \\ \cline{2-5} 
                           & GMM-VRD             & 0.721 (0.013)          & 0.724 (0.015)          & -                      \\ \cline{2-5} 
                           & Dynse               & 0.582 (0.076)          & 0.586 (0.072)          & -                      \\ \cline{2-5} 
                           & IGMM-CD             & 0.497 (0.081)          & 0.578 (0.077)          & -                      \\ \hline
\multirow{12}{*}{Virtual5} & HAT-DDM             & 0.700 (0.043)          & 0.715 (0.041)          & 0.663 (0.043)          \\ \cline{2-5} 
                           & HAT-EDDM            & 0.466 (0.016)          & 0.478 (0.017)          & 0.229 (0.014)          \\ \cline{2-5} 
                           & HAT-ADWIN           & 0.532 (0.053)          & 0.642 (0.054)          & 0.338 (0.063)          \\ \cline{2-5} 
                           & AWE                 & \textbf{0.819 (0.003)} & \textbf{0.937 (0.002)} & \textbf{0.784 (0.004)} \\ \cline{2-5} 
                           & LevBag              & 0.790 (0.004)          & 0.900 (0.002)          & 0.689 (0.003)          \\ \cline{2-5} 
                           & OzaAD               & 0.709 (0.006)          & 0.852 (0.013)          & 0.743 (0.012)          \\ \cline{2-5} 
                           & OzaAS               & 0.704 (0.003)          & 0.804 (0.003)          & 0.696 (0.008)          \\ \cline{2-5} 
                           & ARF                 & 0.803 (0.004)          & 0.882 (0.008)          & 0.827 (0.006)          \\ \cline{2-5} 
                           & OGMMF-VRD           & 0.798 (0.010)          & 0.901 (0.009)          & 0.775 (0.013)          \\ \cline{2-5} 
                           & GMM-VRD             & 0.796 (0.009)          & 0.901 (0.005)          & 0.738 (0.012)          \\ \cline{2-5} 
                           & Dynse               & 0.778 (0.007)          & 0.875 (0.008)          & 0.710 (0.016)          \\ \cline{2-5} 
                           & IGMM-CD             & 0.694 (0.007)          & 0.884 (0.004)          & 0.676 (0.011)          \\ \hline
\multirow{12}{*}{Virtual9} & HAT-DDM             & 0.664 (0.023)          & 0.712 (0.050)          & 0.748 (0.027)          \\ \cline{2-5} 
                           & HAT-EDDM            & 0.380 (0.016)          & 0.485 (0.017)          & 0.360 (0.015)          \\ \cline{2-5} 
                           & HAT-ADWIN           & 0.363 (0.070)          & 0.610 (0.064)          & 0.493 (0.067)          \\ \cline{2-5} 
                           & AWE                 & \textbf{0.833 (0.003)} & \textbf{0.942 (0.002)} & 0.827 (0.003)          \\ \cline{2-5} 
                           & LevBag              & 0.760 (0.013)          & 0.873 (0.010)          & 0.794 (0.006)          \\ \cline{2-5} 
                           & OzaAD               & 0.733 (0.011)          & 0.848 (0.015)          & 0.710 (0.034)          \\ \cline{2-5} 
                           & OzaAS               & 0.710 (0.004)          & 0.784 (0.003)          & 0.682 (0.004)          \\ \cline{2-5} 
                           & ARF                 & 0.806 (0.007)          & 0.889 (0.008)          & \textbf{0.836 (0.005)} \\ \cline{2-5} 
                           & IGMM-CD             & 0.760 (0.017)          & 0.889 (0.008)          & 0.756 (0.011)          \\ \cline{2-5} 
                           & OGMMF-VRD           & 0.814 (0.011)          & 0.922 (0.009)          & 0.834 (0.012)          \\ \cline{2-5} 
                           & GMM-VRD             & 0.810 (0.013)          & 0.913 (0.004)          & 0.807 (0.010)          \\ \cline{2-5} 
                           & Dynse               & 0.773 (0.014)          & 0.866 (0.006)          & 0.772 (0.010)          \\ \hline
\multirow{12}{*}{SEA}      & HAT-DDM             & 0.880 (0.019)          & 0.509 (0.011)          & -                      \\ \cline{2-5} 
                           & HAT-EDDM            & 0.584 (0.025)          & 0.422 (0.026)          & -                      \\ \cline{2-5} 
                           & HAT-ADWIN           & \textbf{0.914 (0.092)} & 0.118 (0.093)          & -                      \\ \cline{2-5} 
                           & AWE                 & 0.878 (0.002)          & 0.828 (0.003)          & -                      \\ \cline{2-5} 
                           & LevBag              & 0.861 (0.003)          & 0.797 (0.005)          & -                      \\ \cline{2-5} 
                           & OzaAD               & 0.815 (0.010)          & 0.796 (0.004)          & -                      \\ \cline{2-5} 
                           & OzaAS               & 0.811 (0.002)          & 0.799 (0.002)          & -                      \\ \cline{2-5} 
                           & ARF                 & 0.838 (0.006)          & \textbf{0.829 (0.010)} & -                      \\ \cline{2-5} 
                           & OGMMF-VRD           & 0.855 (0.007)          & 0.826 (0.008)          & -                      \\ \cline{2-5} 
                           & GMM-VRD             & 0.797 (0.010)          & 0.761 (0.015)          & -                      \\ \cline{2-5} 
                           & Dynse               & 0.701 (0.018)          & 0.750 (0.019)          & -                      \\ \cline{2-5} 
                           & IGMM-CD             & 0.740 (0.011)          & 0.543 (0.015)          & -                      \\ \hline
\multirow{12}{*}{SEARec}   & HAT-DDM             & 0.735 (0.064)          & 0.694 (0.068)          & -                      \\ \cline{2-5} 
                           & HAT-EDDM            & 0.519 (0.017)          & 0.484 (0.017)          & -                      \\ \cline{2-5} 
                           & HAT-ADWIN           & 0.660 (0.133)          & 0.396 (0.153)          & -                      \\ \cline{2-5} 
                           & AWE                 & \textbf{0.876 (0.002)} & \textbf{0.835 (0.002)} & -                      \\ \cline{2-5} 
                           & LevBag              & 0.868 (0.003)          & 0.793 (0.004)          & -                      \\ \cline{2-5} 
                           & OzaAD               & 0.819 (0.004)          & 0.802 (0.004)          & -                      \\ \cline{2-5} 
                           & OzaAS               & 0.821 (0.001)          & 0.798 (0.001)          & -                      \\ \cline{2-5} 
                           & ARF                 & 0.836 (0.006)          & 0.822 (0.009)          & -                      \\ \cline{2-5} 
                           & OGMMF-VRD           & 0.853 (0.005)          & 0.828 (0.007)          & -                      \\ \cline{2-5} 
                           & GMM-VRD             & 0.790 (0.013)          & 0.764 (0.010)          & -                      \\ \cline{2-5} 
                           & Dynse               & 0.734 (0.007)          & 0.760 (0.008)          & -                      \\ \cline{2-5} 
                           & IGMM-CD             & 0.744 (0.009)          & 0.538 (0.013)          & -                      \\ \hline
\multirow{12}{*}{NOAA}     & HAT-DDM             & 0.230 (0.053)          & \textbf{0.925 (0.020)} & -                      \\ \cline{2-5} 
                           & HAT-EDDM            & 0.016 (0.006)          & 0.992 (0.003)          & -                      \\ \cline{2-5} 
                           & HAT-ADWIN           & 0.148 (0.056)          & 0.950 (0.020)          & -                      \\ \cline{2-5} 
                           & AWE                 & 0.569 (0.015)          & 0.821 (0.010)          & -                      \\ \cline{2-5} 
                           & LevBag              & 0.705 (0.013)          & 0.712 (0.010)          & -                      \\ \cline{2-5} 
                           & OzaAD               & \textbf{0.738 (0.036)} & 0.630 (0.037)          & -                      \\ \cline{2-5} 
                           & OzaAS               & 0.613 (0.066)          & 0.721 (0.056)          & -                      \\ \cline{2-5} 
                           & ARF                 & 0.478 (0.015)          & 0.897 (0.004)          & -                      \\ \cline{2-5} 
                           & OGMMF-VRD           & 0.603 (0.009)          & 0.806 (0.006)          & -                      \\ \cline{2-5} 
                           & GMM-VRD             & 0.628 (0.035)          & 0.735 (0.030)          & -                      \\ \cline{2-5} 
                           & Dynse               & 0.462 (0.006)          & 0.750 (0.004)          & -                      \\ \cline{2-5} 
                           & IGMM-CD             & 0.573 (0.007)          & 0.619 (0.010)          & -                      \\ \hline
\multirow{12}{*}{ELEC}     & HAT-DDM             & 0.494 (0.033)          & 0.841 (0.018)          & -                      \\ \cline{2-5} 
                           & HAT-EDDM            & 0.439 (0.016)          & 0.843 (0.008)          & -                      \\ \cline{2-5} 
                           & HAT-ADWIN           & 0.489 (0.023)          & 0.844 (0.014)          & -                      \\ \cline{2-5} 
                           & AWE                 & 0.589 (0.005)          & 0.798 (0.004)          & -                      \\ \cline{2-5} 
                           & LevBag              & 0.656 (0.007)          & 0.835 (0.006)          & -                      \\ \cline{2-5} 
                           & OzaAD               & 0.563 (0.025)          & 0.807 (0.021)          & -                      \\ \cline{2-5} 
                           & OzaAS               & 0.357 (0.007)          & \textbf{0.907 (0.004)} & -                      \\ \cline{2-5} 
                           & ARF                 & 0.705 (0.007)          & 0.857 (0.006)          & -                      \\ \cline{2-5} 
                           & OGMMF-VRD           & 0.659 (0.011)          & 0.807 (0.008)          & -                      \\ \cline{2-5} 
                           & GMM-VRD             & 0.606 (0.020)          & 0.740 (0.017)          & -                      \\ \cline{2-5} 
                           & Dynse               & 0.508 (0.004)          & 0.752 (0.002)          & -                      \\ \cline{2-5} 
                           & IGMM-CD             & \textbf{0.793 (0.002)} & 0.777 (0.002)          & -                      \\ \hline
\multirow{12}{*}{PAKDD}    & HAT-DDM             & 0.996 (0.001)          & 0.005 (0.002)          & -                      \\ \cline{2-5} 
                           & HAT-EDDM            & \textbf{1.000 (0.000)} & 0.000 (0.000)          & -                      \\ \cline{2-5} 
                           & HAT-ADWIN           & 0.996 (0.002)          & 0.005 (0.002)          & -                      \\ \cline{2-5} 
                           & AWE                 & 0.656 (0.086)          & 0.476 (0.098)          & -                      \\ \cline{2-5} 
                           & LevBag              & 0.826 (0.014)          & 0.245 (0.020)          & -                      \\ \cline{2-5} 
                           & OzaAD               & 0.299 (0.048)          & \textbf{0.788 (0.041)} & -                      \\ \cline{2-5} 
                           & OzaAS               & 0.537 (0.094)          & 0.594 (0.083)          & -                      \\ \cline{2-5} 
                           & ARF                 & 0.982 (0.001)          & 0.037 (0.003)          & -                      \\ \cline{2-5} 
                           & OGMMF-VRD           & 0.948 (0.002)          & 0.061 (0.001)          & -                      \\ \cline{2-5} 
                           & GMM-VRD             & 0.830 (0.015)          & 0.212 (0.017)          & -                      \\ \cline{2-5} 
                           & Dynse               & 0.417 (0.002)          & 0.644 (0.002)          & -                      \\ \cline{2-5} 
                           & IGMM-CD             & 0.376 (0.001)          & 0.601 (0.001)          & -                      \\ \hline

\end{longtable}